\documentclass{article}
\usepackage{arxiv}
\usepackage[authoryear]{natbib}
\usepackage[utf8]{inputenc} % allow utf-8 input
\usepackage[T1]{fontenc}    % use 8-bit T1 fonts
\usepackage{hyperref}       % hyperlinks
\usepackage{url}            % simple URL typesetting
\usepackage{booktabs}       % professional-quality tables
\usepackage{amsfonts}       % blackboard math symbols
\usepackage{nicefrac}       % compact symbols for 1/2, etc.
\usepackage{microtype}      % microtypography
\usepackage{graphicx}
\usepackage{subcaption}
\usepackage{tcolorbox}
\usepackage{multirow}
\usepackage{diagbox}
\usepackage[normalem]{ulem}
\usepackage{doi}
\usepackage{placeins} % 引入 placeins 宏包
\usepackage{enumitem}
\usepackage{amsmath} 
\usepackage{tabularx} % 在导言区导入 tabularx 包
\usepackage{hyperref}
\usepackage{makecell}
\usepackage{tcolorbox}
\tcbuselibrary{listings}

\newtcblisting{mylatexcode}{
	listing only,
	colback=gray!5!white,  % 背景色
	colframe=gray!50!black, % 边框颜色
	listing options={basicstyle=\ttfamily\small}
}

\title{Every FLOP Counts: Scaling a 300B Mixture-of-Experts LING LLM without premium GPUs}
\author{Ling Team, AI@Ant Group}
\begin{document}
	\maketitle
	\thispagestyle{firstpage}
	\begin{abstract}
		In this technical report, we tackle the challenges of training large-scale Mixture of Experts (MoE) models, focusing on overcoming cost inefficiency and resource limitations prevalent in such systems. To address these issues, we present two differently sized MoE large language models (LLMs), namely Ling-Lite and Ling-Plus (referred to as "Bailing" in Chinese, spelled Bǎilíng in Pinyin). Ling-Lite contains 16.8 billion parameters with 2.75 billion activated parameters, while Ling-Plus boasts 290 billion parameters with 28.8 billion activated parameters. Both models exhibit comparable performance to leading industry benchmarks. This report offers actionable insights to improve the efficiency and accessibility of AI development in resource-constrained settings, promoting more scalable and sustainable technologies. Specifically, to reduce training costs for large-scale MoE models, we propose innovative methods for (1) optimization of model architecture and training processes, (2) refinement of training anomaly handling, and (3) enhancement of model evaluation efficiency. Additionally, leveraging high-quality data generated from knowledge graphs, our models demonstrate superior capabilities in tool use compared to other models. Ultimately, our experimental findings demonstrate that a 300B MoE LLM can be effectively trained on lower-performance devices while achieving comparable performance to models of a similar scale, including dense and MoE models. Compared to high-performance devices, utilizing a lower-specification hardware system during the pre-training phase demonstrates significant cost savings, reducing computing costs by approximately 20\%. The models can be accessed at \href{https://huggingface.co/inclusionAI}{https://huggingface.co/inclusionAI}.

		\FloatBarrier % 插入屏障，强制处理之前的浮动体
		\begin{figure}[]
			\begin{center}	
				\includegraphics[width=1.0\textwidth]{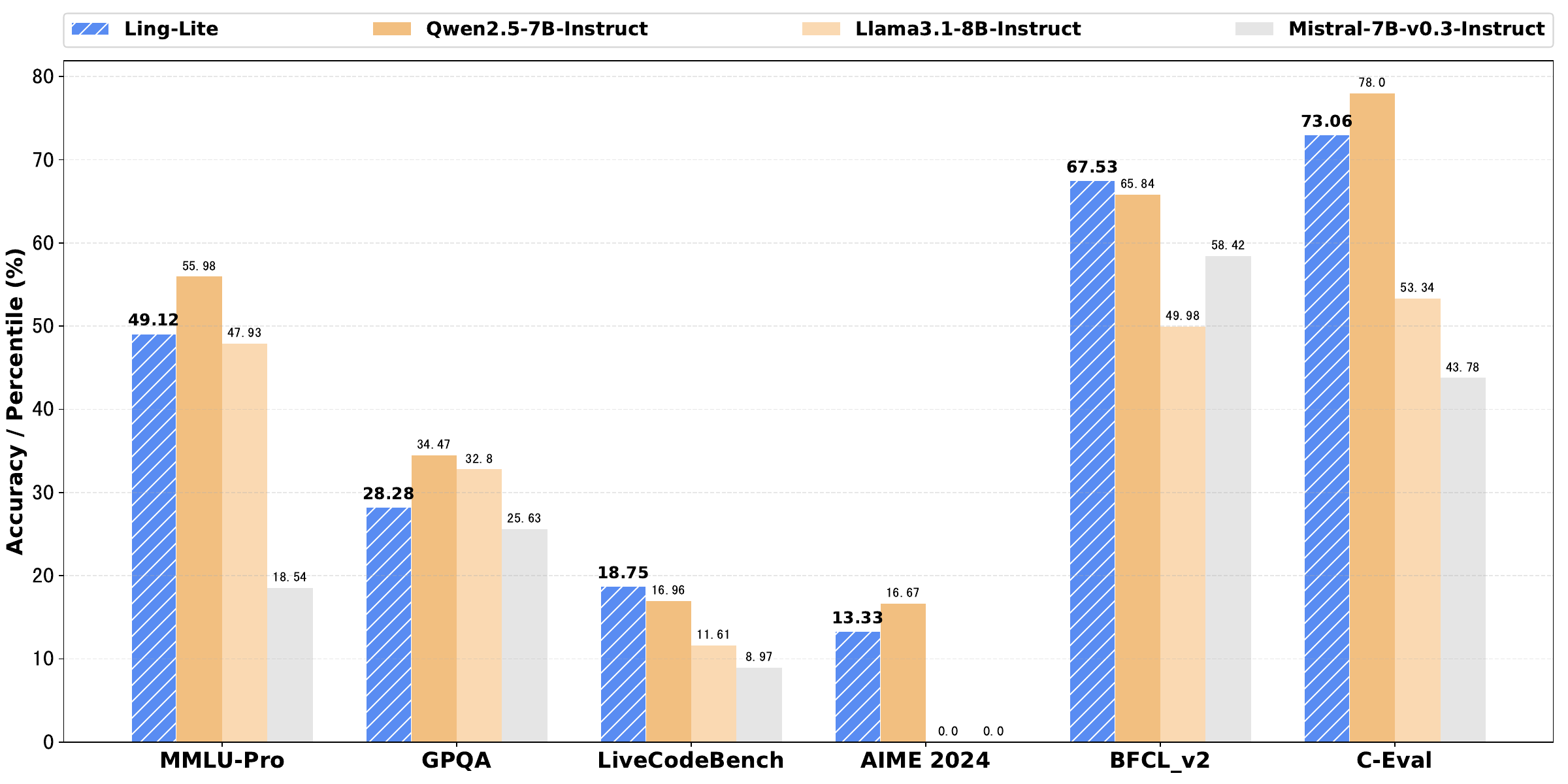}
				\caption{Ling-Lite performance. }
				\label{fig:linglite}
			\end{center}
		\end{figure}
		\begin{figure}[]
			\begin{center}	
				\includegraphics[width=1.0\textwidth]{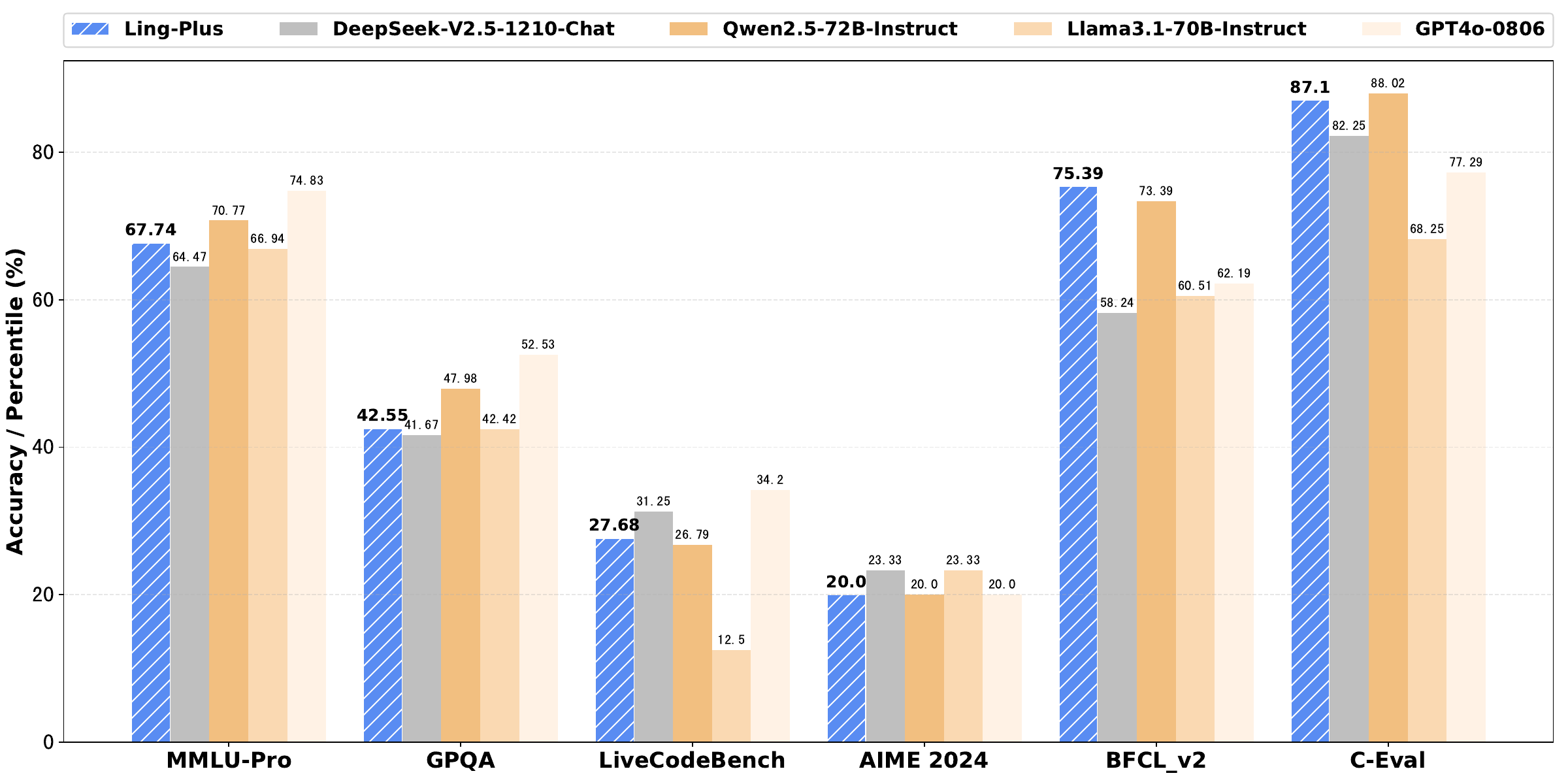}
				\caption{Ling-Plus performance.}
				\label{fig:lingplus}
			\end{center}
		\end{figure}
	\end{abstract}
	
	\section{Introduction}
\subsection{Background and Motivation}
In recent years, the rapid development of LLMs \cite{openai2024gpt4technicalreport,geminiteam2024gemini15unlockingmultimodal,claude2024claude3,qwen2025qwen25technicalreport,deepseekai2025deepseekv3technicalreport} has sparked widespread discussions across academia and industry regarding Artificial General Intelligence (AGI). While dense models have achieved remarkable progress, MoE models, such as the DeepSeek series \cite{deepseekai2024deepseekllmscalingopensource,deepseekai2024deepseekv2strongeconomicalefficient,deepseekai2025deepseekv3technicalreport}, the Qwen series \cite{bai2023qwentechnicalreport,yang2024qwen2technicalreport,qwen2025qwen25technicalreport}, and the MiniMax-01 series \cite{minimax2025minimax01scalingfoundationmodels}, have demonstrated outstanding performance, even surpassing traditional dense models in certain specific tasks. However, the training of MoE models typically relies on high-performance computing resources (e.g., advanced AI accelerator like H100 and H800), and their prohibitively high costs have limited broader adoption in resource-constrained environments. This study proposes innovative training strategies to enable efficient LLM training under restricted resources and budget constraints, thereby advancing the inclusive development of AI technologies. To provide the industry with a novel approach to model training in resource-constrained scenarios and to inspire the development of more innovative solutions, this report introduces our open-source MoE models, Ling-Lite (with a total parameter count of 16.8B and an activation parameter count of 2.75B) and Ling-Plus (with a total parameter count of 290B and an activation parameter count of 28.8B), focusing on their exploration and optimization practices.

\subsection{Computing Environment for Model Training}
The availability of computational resources is a critical determinant in the development of LLMs, particularly in the context of the increasingly popular MoE architecture. Recent State-of-the-art MoE models rely heavily on high-performance AI accelerators (e.g., H100 and H800) for training, yet the supply of such resources has remained constrained in recent years. Similar to the analysis on the imbalance of day-night inference load in the DeepSeek's open-source files \cite{deepseek2025opensourceweek}, in the commercial deployment of AI services, these high-performance resources are also in high demand during peak usage periods to ensure service quality. As a result, many LLM research organizations face persistent shortages of high-performance AI accelerators.
\begin{table}[t]
\centering
\caption{Characteristics of different AI accelerators (listed in descending order of availability).}
\vspace{3mm}
\label{tab:computational_resources}
\resizebox{0.7\linewidth}{!}{%
\begin{tabular}{l|c|c|c|c}
\toprule
\textbf{Device} & \textbf{Peak FLOPS (T)} & \textbf{Memory (GB)} & \textbf{Fair Cost per Hour (RMB)} & \textbf{Support FP8}\\
    \midrule
    A& 370 & 64 & 7 & $\times$\\ 
    \midrule
    B& 120& 96 & 4.5 & $\times$\\ 
    \midrule
    C & 312& 80  & 10 & $\times$ \\  
    \midrule
    D& 989& 80& 27.5 & $\surd$ \\ 
    \midrule
    E & 147 & 96 & 5.64 & $\surd$ \\ 
    \bottomrule
\end{tabular}
}
\end{table}
In comparison, lower-performance accelerators are more widely available and maybe cost-effective on a per-unit basis. This discrepancy highlights the need for a technical framework that enables seamless switching between heterogeneous computing units and distributed clusters for training and inference. Such a system could alleviate the supply-demand imbalance and reduce overall training costs. The training of our Ling models utilized the computational resources in Table \ref{tab:computational_resources}.

From an economic efficiency perspective, these solutions reduce unit compute costs. However, the heterogeneous nature of device architectures (e.g., DSA and GPGPU) and the geographical dispersion of clusters introduce significant technical challenges, primarily in the following three aspects.

\begin{itemize}
    \item \textbf{Cross-cluster and cross-device compatibility.} As described in the open-source project FlagScale \cite{flagscale2025github}, heterogeneous hardware environments often exhibit discrepancies in the implementation of low-level computational and communication operators and high-level distributed training frameworks. These challenges are particularly evident in training advanced architectures like MoE, where operators such as \texttt{group\_gemm}, \texttt{permute/unpermute}, and \texttt{all2all}, and distributed strategies like expert parallelism may be missing or perform inconsistently across platforms. Ensuring training accuracy and portability requires: (1) \textit{collaboration with hardware vendors} to standardize low-level operators, ensuring computational and communication consistency, (2) \textit{development of cross-platform compatibility layers} to support seamless integration across distributed training frameworks, and (3) \textit{implementation of efficient debugging mechanisms} for identifying and resolving issues in complex and heterogeneous environments.
    \item \textbf{Reliability of cross-cluster resource synchronization.} LLM training, especially for ultra-large MoE models, requires managing massive datasets and checkpoint backups that can reach petabyte (PB) scales. Seamless task migration across clusters relies on: (1) \textit{achieving low-latency, consistent synchronization} of data resources across clusters and (2) \textit{enabling flexible, high-speed management and I/O for training artifacts}, such as model checkpoints, across distributed clusters.
    \item \textbf{Cost-performance optimization.} Balancing cost efficiency and model performance is a core objective. Achieving this requires: (1) \textit{optimization of hardware resource allocation and scheduling}, (2) \textit{trade-offs between computational efficiency and training precision}, and (3) \textit{rational design and scaling} of model architectures to ensure cost-effective performance.
\end{itemize}

\subsection{Optimization for Model Training}
To address the above-mentioned technical challenges posed by limited computational resources, we implement a series of systematic optimization strategies to balance resource cost and model performance. These strategies are outlined as follows.
\begin{itemize}
    \item \textbf{Optimization of model architecture and training strategies.} To enable efficient deployment on resource-constrained platforms, we adpot the following three strategies. (1) \textit{Model architecture optimization}: Based on the comprehensive analysis of scaling laws for dense and MoE models, we can choose the best-matching architecture for the available computational resource. (2) \textit{Training framework optimization}: For heterogeneous computing platforms, we integrate multiple training frameworks into a unified distributed deep learning framework, i.e., our open-source project, DLRover \cite{dlrover2023github}. Additionally, to leverage the specific characteristics of various platforms, we develop a lightweight debugging tool, XPUTimer, which facilitates rapid and cost-effective task performance analysis while achieving a 90\% reduction in memory usage. Furthermore, we implement a platform-agnostic asynchronous training strategy, namely EDiT (Elastic Distributed Training), which enhances the training efficiency, under various configurations, the training time can be reduced by up to 66.1\%. (3) \textit{Storage optimization}: Techniques such as device multi-tenancy and file system in user space (FUSE) are applied to achieve high performance and multi-cluster adaptability for large-scale training. Collaborative design of storage and training processes enhances I/O efficiency in MoE scenarios, reducing time overhead by 50\%.
    \item \textbf{Refinement of training anomaly handling.} To address hardware errors and loss anomalies in large-scale training, we develop a robust anomaly-handling mechanism as follows. (1) \textit{Multi-level anomaly detection system:} To detect anomalies throughout the training process, we establish a real-time monitoring system. (2) \textit{Automated checkpoint recovery:} To minimize the impact of anomalies on training progression, we implement an automated recovery mechanism.
    \item \textbf{Enhancement of model evaluation efficiency.} To optimize monitoring of cross-cluster model training, we attempt to improve the evaluation benchmarks and frameworks as follows. (1) \textit{Comprehensive evaluation dataset:} To mitigate initial model underperformance and improve stability, we construct some domain-specific evaluation datasets and optimize the corresponding prediction strategies and prompting templates. (2) \textit{Efficient evaluation system:} Based on our self-innovate offline inference framework, i.e., Flood, we develop a scalable system for cross-cluster evaluations with consistent results, achieving an average deviation of less than 0.5\%. (3) \textit{Automated analysis system:} To provide real-time feedback to adjust training strategies, we develop an automated system to correlate evaluation results with model performance and datasets.
    \item \textbf{Improvement of tool use capability.} To enhance the tool use ability of large models, we focus on the following two key aspects. (1) \textit{High-quality data synthesis:} To efficiently generate high-quality, scalable, and diverse tool-use data, we leverage knowledge graph technology and generalized calling instructions to extract diverse and complex function chains and thus enhance the applicability of Ling models across various real-world scenarios. (2) \textit{Adaptive tool learning:} By leveraging learning strategies such as rejection sampling and error correction, we develop self-reflective multi-agent interactive dialogues to enhance the adaptive tool use capability of the Ling model.
\end{itemize}

Based on the above-mentioned technical optimizations, we develop and open-source the Ling series of MoE models, which achieves a balanced trade-off between resource cost and model performance. From the perspective of resource efficiency, Ling-Plus serves as an illustrative example, with pre-training conducted on 9 trillion tokens across five distinct hardware configurations (as detailed in Table \ref{tab:computational_resources}). Training 1 trillion tokens using the high-performance hardware configuration (device D) incurs an estimated cost of approximately 6.35 million RMB. In contrast, utilizing a lower-specification hardware system reduces the cost to around 5.08 million RMB, representing a cost savings of nearly 20\%. These results demonstrate the feasibility of training state-of-the-art (SOTA) large-scale MoE models on less powerful hardware, enabling a more flexible and cost-effective approach to foundational model development with respect to computing resource selection.

We evaluated our Ling models on a comprehensive array of benchmarks. With similar parameter sizes, our Ling models trained under limited resources and budget constraints deliver comparable performance to existing open-source models, particularly in the ability of tool use (see the evaluation results in Figures \ref{fig:linglite} and \ref{fig:lingplus}).

\subsection{Challenges and Lessons Learned}
Despite the above-mentioned contributions, the process of transitioning training tasks across different accelerators continues to pose significant challenges. Throughout the training process, several issues were identified, which are outlined below along with the key insights derived from addressing them:
\begin{itemize}
    \item \textbf{Training stability.} During the training of ultra-large-scale models, both hardware-related factors and seemingly minor modifications to the network structure can substantially influence the stability and convergence of the models. In particular, challenges such as loss divergence, loss spikes, and expert load imbalance were observed. These issues, along with the strategies employed to address them, are thoroughly discussed in Section \ref{bitter_lessons}.
    \item \textbf{Cross-platform alignment.} When migrating training workflows across different hardware environments, the upper-layer framework provides abstraction and ensures the accuracy of basic operations. However, minor precision errors can accumulate throughout the course of large-scale training. Over time, these seemingly negligible discrepancies can lead to significant variations in outcomes across different hardware configurations.
\end{itemize}

In the following sections, we will introduce our Ling models in the sequence of Infrastructure (Section \ref{section_infrastruction}), Pre-Training (Section \ref{section_pre_training}), and Post-Training (Section \ref{section_post_training}). Finally, we will present the model's performance on the evaluation benchmarks in Section \ref{section_results}, along with some of the lessons learned throughout the process in Section \ref{bitter_lessons}.
	\section{Infrastructure, Scaling, and Efficiency} \label{section_infrastruction}
In response to the growing demand for high-performance accelerators required for training large-scale models, expanding computational capacity through the integration of additional hardware has become an essential strategy. To address this challenge , we leverage our open-source project DLRover (Distributed Deep Learning Training System) to optimize and seamlessly migrate computing workloads to proprietary hardware. With the help of this framework, it is very easy to launch training frameworks on different platforms, including DeepSpeed \cite{song2023deepspeed4scienceinitiativeenablinglargescale}, Megatron-LM \cite{shoeybi2020megatronlmtrainingmultibillionparameter}, and Megatron vendor version. To meet the need for lightweight performance monitoring and fault diagnosis, DLRover incorporates the XPUTimer \cite{cui2025xputimer}, a minimalistic runtime performance analysis framework. 
Furthermore, to mitigate performance decline in large-scale heterogeneous distributed training environments, the EDiT~\cite{cheng2025edit} method has been adopted, which is an efficient asynchronous training approach tailored for LLMs.
In addition to computational efficiency, I/O also has a significant impact on overall performance. To address this, we have developed PCache and a cross-cluster synchronization solution, further enhancing the overall training efficiency.
Lastly, to improve data synthesis efficiency and accelerate the evaluation process, a high-performance offline inference framework, named Flood, has been introduced.

\subsection{Lightweight Profiler}
\begin{figure}[t]
\begin{center}	
\includegraphics[width=0.7\textwidth]{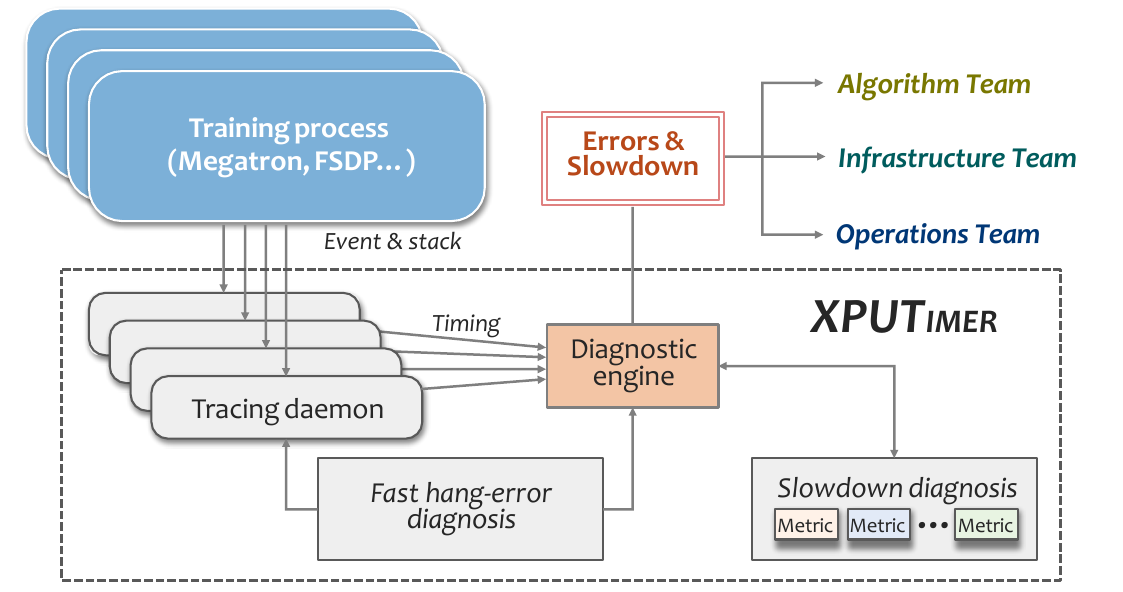}
\caption{The general structure of XPUTimer.}
\label{xpu_timer}
\end{center}
\end{figure}

To address performance bottlenecks and hidden inefficiencies in distributed training of large-scale models, we propose a lightweight analytical tool, referred to as XPUTimer (see Figure \ref{xpu_timer}). XPUTimer has been integrated into our open-source DLRover system, enabling real-time diagnostic capabilities across the entire training stack. This tool also facilitates the retrieval of status information from diverse training environments. As illustrated in Figure \ref{xpu_timer}, XPUTimer comprises two primary components: (1) \textit{lightweight selective tracing} and (2) \textit{diagnostic engine}. The selective tracing mechanism is designed to monitor critical training code segments while incurring minimal overhead. The diagnostic engine, in turn, leverages the real-time data collected by a tracking daemon to rapidly pinpoint the root causes of training anomalies.
\subsubsection{Lightweight Selective Tracing}
\begin{figure}[t]
\begin{center}	
\includegraphics[width=0.7\textwidth]{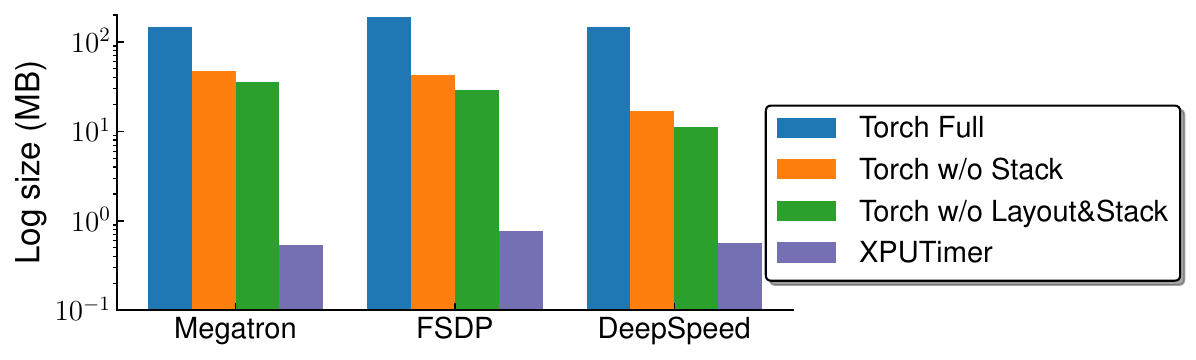}
\caption{The memory usage comparisons between XPUTimer and other methods.}
\label{fig:xpumemory_baseline}
\vspace{4mm}
\includegraphics[width=0.75\textwidth]{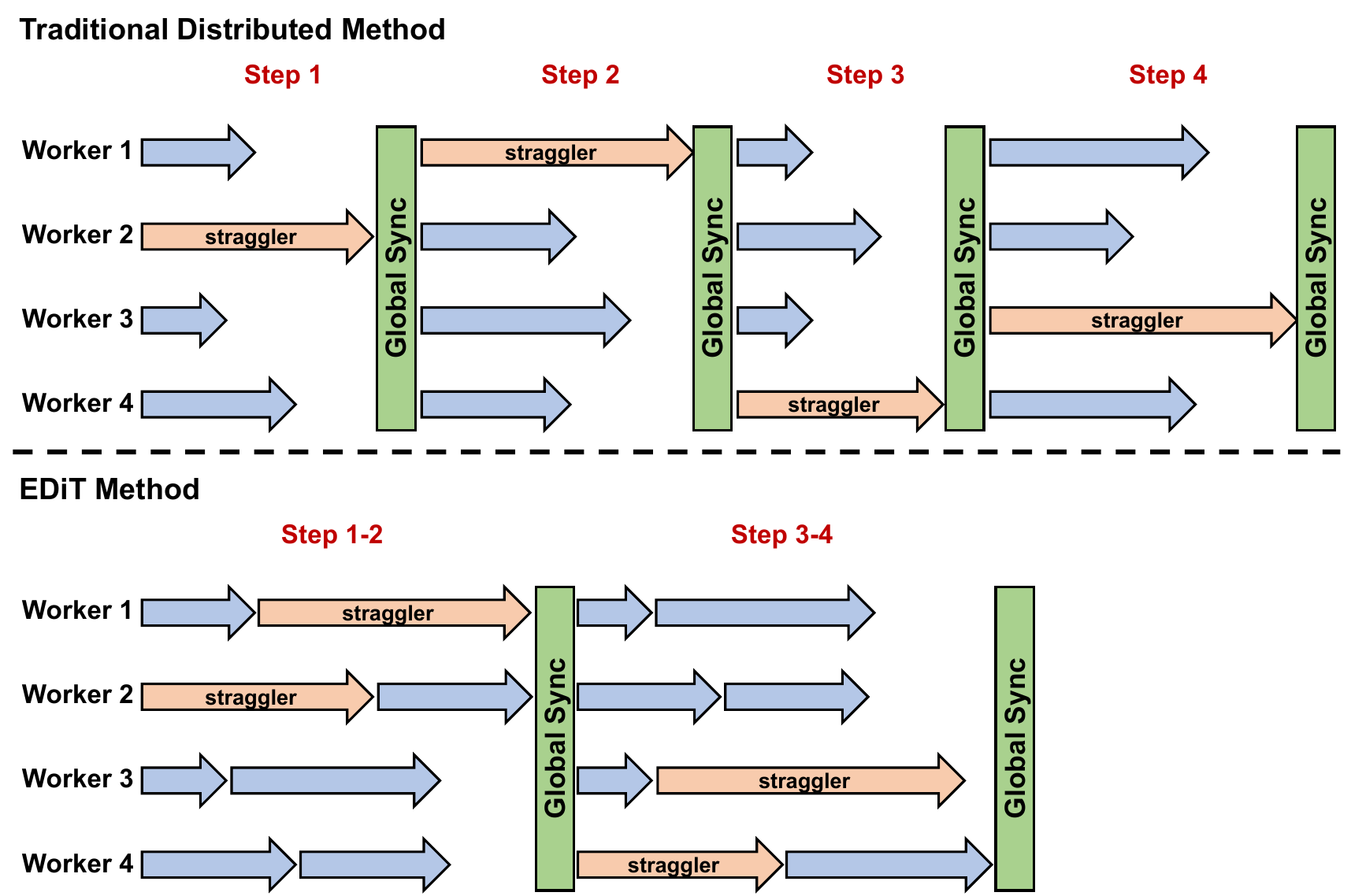}
\caption{The comparisons of traditional distributed method and EDiT method.}
\label{fig:EDiT_comparisons}
\end{center}
\end{figure}

The lightweight selective tracing mechanism is designed to capture and log critical events selectively, ensuring that sufficient diagnostic information is collected without incurring the substantial memory and computational overhead of full-scale monitoring. The key features of this mechanism can be summarized as follows:
\begin{itemize}
    \item \textbf{Error interception.} To identify high-level operations that may introduce performance bottlenecks or errors, we implemented Python-layer interception that allows dynamic configuration of APIs for monitoring (e.g., garbage collection, synchronization, and data loading). This is achieved by modifying environment variables such as TRACED\_PYTHON\_API. In addition, we designed a framework-agnostic kernel monitoring mechanism using C++/CUDA-level interception. This approach enables the tracking of computation kernels (e.g., cuBLAS and Flash Attention), communication kernels (e.g., NCCL operations), and custom operators via an explicit registration interface.
    \item \textbf{Interference avoidance.} XPUTimer minimizes its impact on the training process by employing asynchronous event management. Specifically, it combines synchronous APIs for timestamp recording with asynchronous kernels of accelerators using CUDA events to monitor execution states. For instance, events are injected following NCCL kernel launches, and their completion is monitored in a background thread. This approach ensures that the diagnostic process does not interfere with the primary training workflow.
    \item \textbf{Low overhead.} XPUTimer is designed to maintain a low-cost diagnostic footprint. To enable asynchronous event management, it employs an optimized architecture that includes: (1) \textit{event pool management} to reuse pre-allocated CUDA events, (2) \textit{asynchronous data processing} via a dedicated background thread for event collection and logging, and (3) \textit{data compression techniques} that record only essential fields, such as timestamps and kernel input layouts. These optimizations lead to significantly reduced log sizes, averaging approximately 1.5 MB per accelerator per training step, which represents an approximate 90\% reduction in memory usage, as illustrated in Figure \ref{fig:xpumemory_baseline}. 
\end{itemize}

\subsubsection{Diagnostic Engine}The diagnostic engine is also a core component of XPUTimer, responsible for analyzing real-time data collected by the tracing daemon to quickly pinpoint the root causes of training anomalies. It tackles the attribution challenges in large-scale distributed training through two key modules: error diagnosis and performance degradation diagnosis. The design of the diagnostic engine is built around two core objectives:
\begin{itemize}
    \item \textbf{Fast attribution.} Through the implementation of a multi-layered diagnostic approach that integrates call stack analysis with in-kernel tracing, the process of error localization is significantly optimized, reducing the time complexity from the conventional $O(logN)$ to $O(1)$.
    \item \textbf{Fine-grained diagnostics.} By combining macro-level metrics (e.g., throughput) with micro-level metrics (e.g., kernel launch latency distribution), it enables anomaly detection across computation, communication, and non-critical operations such as data loading.
\end{itemize}

\subsection{High-Performance Training Strategy}
\begin{figure}[t]
\begin{center}	
\includegraphics[width=1.0\textwidth]{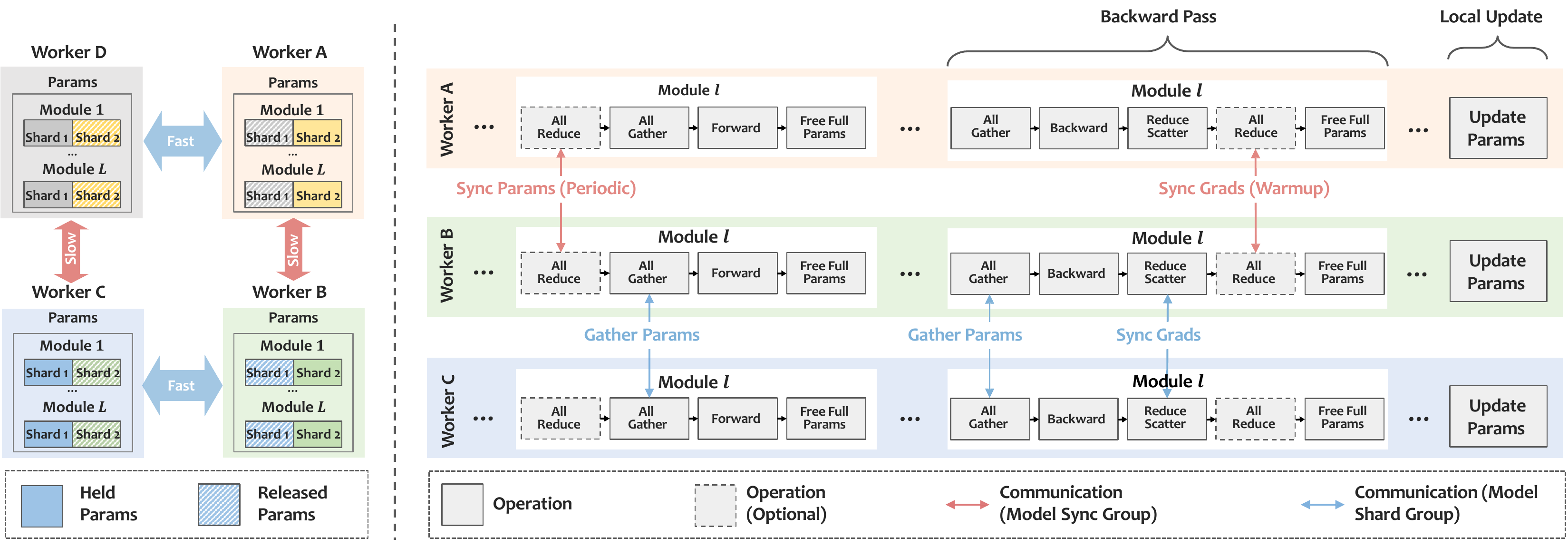}
\caption{The schematic illustration of the EDiT method with 4 workers as an example.}
\label{fig:improved_training_process}
\vspace{4mm}
\includegraphics[width=0.95\textwidth]{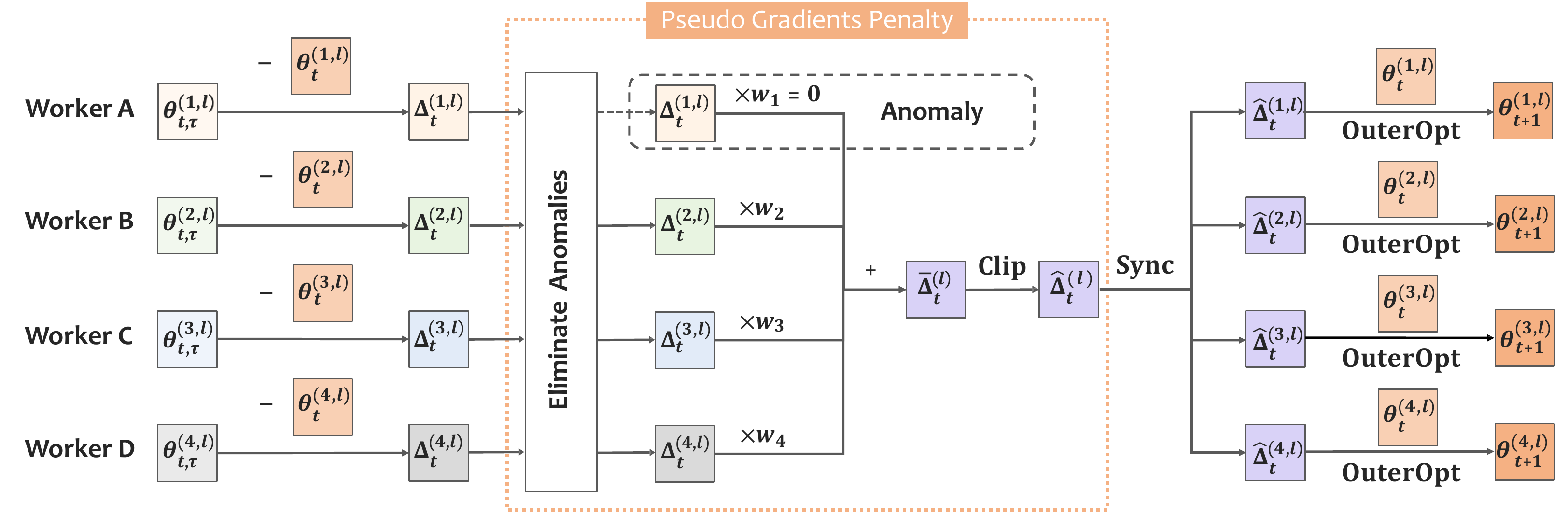}
\caption{The illustration of pseudo gradient penalty strategy in EDiT method.}
\label{fig:pseduo_gradients_penalty}
\end{center}
\end{figure}

With the explosive growth of model size and training data volume, distributed training methods have become critical for efficient training. However, traditional synchronous distributed training methods (e.g., All-Reduce) face the following challenges: (1) high communication overhead, (2) straggler problem, (3) difficulty in elastic training, and (4) sensitivity to data noise. To address these problems, we adopted EDiT method~\cite{cheng2025edit}, which combines a tailored Local SGD (Stochastic Gradient Descent) approach with model sharding techniques to enhance large-scale training efficiency. The comparisons of EDiT and traditional distributed method are illustrated in Figure~\ref{fig:EDiT_comparisons}, and the pipeline of EDiT is illustrated in Figure~\ref{fig:improved_training_process}. The characteristics of EDiT can be summarized as follows:

\begin{itemize}
    \item \textbf{Layer-wise synchronization.} Different from other Local SGD-based methods, EDiT synchronizes parameters layer by layer during forward propagation, significantly reducing the volume of data communicated in a single operation. With the prefetch method, the communication and computation are further overlapped, minimizing idle time and improving overall efficiency.

    \item \textbf{Pseudo gradient penalty.} EDiT employs a pseudo gradient penalty strategy to suppress the loss spikes caused by diverse large-scale corpus and leverages the differences among workers to improve model performance, which is illustrated in Figure~\ref{fig:pseduo_gradients_penalty}. This strategy consists of anomaly elimination, weighted averaging, and gradient clipping. 
    \begin{enumerate}[label=(\arabic*)]
    \item \textit{Anomaly elimination.} The pseudo gradients of each worker are tracked using exponential moving average to detect anomalous workers, which are subsequently excluded from the synchronization process.
    \item \textit{Weighted averaging.} Contributions from workers are weighted based on their pseudo gradient norms, effectively reducing the influence of noisy or outlier gradients on overall model updates.
    \item \textit{Gradient clipping.} A predefined threshold is applied to clip overly large pseudo gradients, ensuring gradient steps remain within a stable range and preventing training instability or divergence.
    \end{enumerate}

    \item \textbf{Time-based synchronization.} Instead of synchronizing after a fixed number of iterations (as in conventional Local SGD-based methods), synchronization can also be triggered based on a time threshold in the EDiT method, enabling faster nodes to perform more local updates before syncing. By decoupling synchronization frequency from iteration counts, EDiT solves the problem of fixed stragglers. Furthermore, this adaptive synchronization mechanism enhances scalability and resilience by dynamically balancing workloads, particularly in heterogeneous environments with diverse hardware capabilities.  
\end{itemize}

\begin{figure}[t]
\begin{center}	
\includegraphics[width=0.7\textwidth]{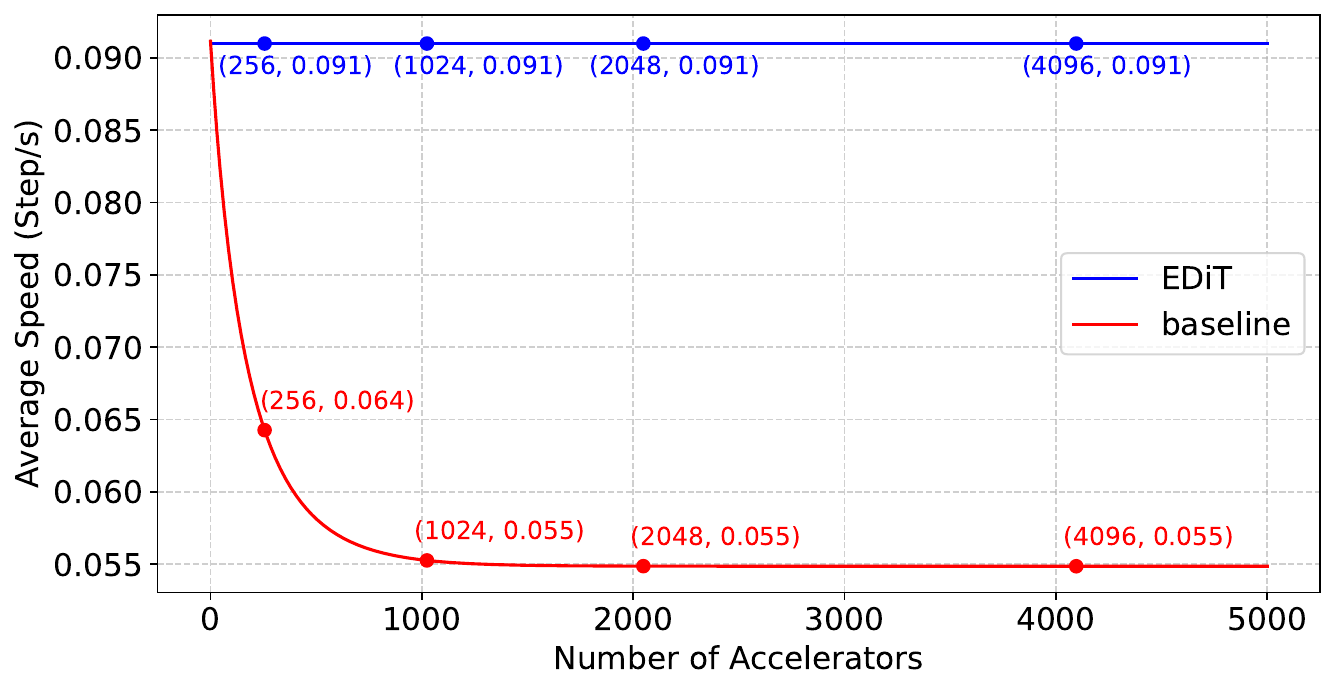}
\caption{The speed comparisons of traditional distributed method and EDiT.}
\label{fig:edit_baseline}
\end{center}
\end{figure}

As shown in Figure~\ref{fig:edit_baseline}, in an ideal environment, as the number of accelerators increases, the minimum speed of the baseline approaches $5.49e^{-2}$ step/s, at which point the speed-up ratio of EDiT would reach $66.1\%$. In practice, however, the average step time of slow steps tends to grow longer as the number of accelerators increases or the model size increases, making the actual acceleration effect of EDiT even more pronounced.

\subsection{Efficient and Highly-Reliable Cross-Cluster Data Synchronization}
The training of modern MoE models requires processing massive datasets, often involving concurrent training and data processing tasks across distributed clusters. This necessitates efficient and reliable access to diverse datasets across clusters, which becomes challenging in cross-cluster environments. To address this problem, we will introduce (1) \textit{a robust distributed storage system} and (2) \textit{an optimized cross-cluster synchronization mechanism} respectively to support the needs of distributed training in the following section.

\subsubsection{Distributed Storage System}
The exponential growth of data and increasing demands for high-performance input/output (I/O) in MoE models have set higher standards for storage system performance. We propose an in-house solution, PCache, has been developed as an all-flash distributed file caching system. PCache is specifically designed to support large-scale internal model training, addressing the performance and scalability requirements of modern distributed training environments. 

\begin{figure}[h]
\begin{center}	
\includegraphics[width=1.0\textwidth]{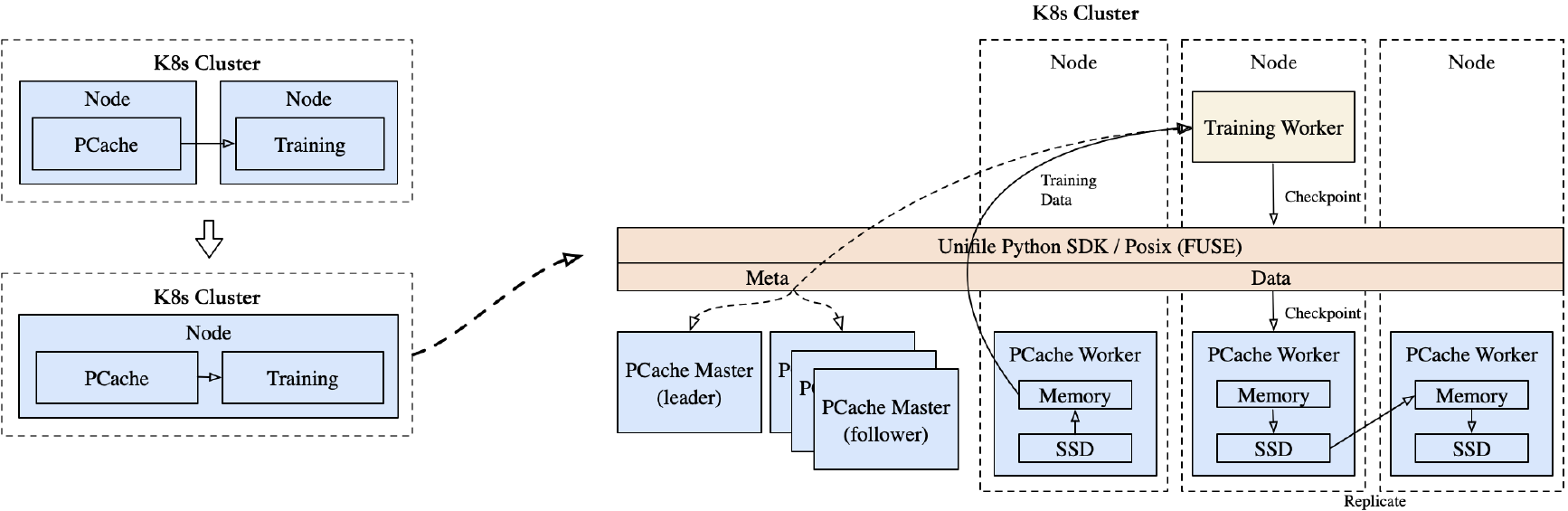}
\caption{The general structure of acclerator multi-tenancy.}
\label{GPU_Multi_Tenancy}
\end{center}
\end{figure}
In contemporary storage architectures, most storage services are deployed as independent clusters. However, in multi-cluster environments, device multi-tenancy has become a crucial requirement. This shift introduces two significant challenges:
\begin{itemize}
    \item \textbf{Dependence on cluster provider storage services.} Each cluster provider offers varied storage capabilities, leading to potential inconsistencies in training performance across heterogeneous cluster environments. 
    \item \textbf{Challenges in building custom storage services.} Independently deployed storage systems face challenges related to hardware compatibility across multiple sites, further compounded by the associated operational and maintenance costs. Networking constraints exacerbate these difficulties. Some cluster providers offer high-performance networks for data transfer, while others do not. 
\end{itemize}
To address these challenges, we have designed a storage service specifically optimized for large-scale model training (see Figure \ref{GPU_Multi_Tenancy}), incorporating the following core features:
\begin{itemize}
    \item \textbf{Broad hardware compatibility and scalability.} The system ensures seamless integration across multi-cluster environments.
    \item \textbf{Cost-performance balance.} It achieves a competitive trade-off between operational costs and performance optimization, ensuring that scalability does not compromise efficiency.
\end{itemize}
Currently, most accelerators are equipped with four or more NVMe SSDs, a configuration that serves as the foundation for PCache’s design. To optimize performance and resource utilization, PCache employs a mixed deployment model, integrating the storage service directly onto computing nodes. This integrated approach minimizes the costs associated with standalone deployments while significantly reducing network latency through localized data processing. 
This ensures that storage throughput grows proportionally with the expansion of computing cluster computational capabilities, providing a scalable and efficient solution for large-scale distributed training systems.
\paragraph{I/O Performance Optimization.}
During the distributed training of MoE models, it is essential to optimize I/O performance to prevent excessive time consumption caused by reading data, checkpoints, and other resources, which can negatively impact training efficiency. The optimization strategies adopted by the PCache system are as follows:
\begin{itemize}
    \item \textbf{File system in user space (FUSE).} In checkpoint read/write scenarios, PCache employs an interception mechanism to eliminate the overhead of multiple switches or data copies between user space and kernel space. By leveraging shared memory (shm), the system accelerates access efficiency for medium to large files.
    \item \textbf{Metadata cache.} A metadata caching strategy combining file-level and data block-level caching significantly enhances client-side read performance, particularly in scenarios involving high volumes of random reads.
    \item \textbf{Worker selection strategy.} This strategy prevents performance degradation caused by cluster bottlenecks or hotspots on overloaded machines.
\end{itemize}

Based on these four optimization strategies, the PCache system achieves the following performance outcomes: 
\begin{itemize}
    \item \textbf{Single client performance.} For scenarios involving single-threaded large file writes, PCache achieves throughput rates of 3–4 GB/s. In multi-threaded scenarios, throughput increases to 20–30 GB/s.
    \item \textbf{Cluster-wide throughput.} Across a 1,000-accelerator cluster, PCache delivers aggregate throughput of 1 TB/s. For clusters with 10,000 accelerators, throughput scales linearly to 8 TB/s.
\end{itemize}
\paragraph{AI Co-Design.}
Megatron, a commonly used framework for training tasks in MoE scenarios, operates with a specific design for checkpoint writing. During the checkpoint writing phase, model and optimizer data are written based on Data Parallel (DP) groups, with the default behavior assigning the responsibility for data aggregation and storage to the rank\_0 device of each DP group. However, this approach can lead to resource contention, as the rank\_0 devices from all DP groups are often concentrated on specific physical nodes. This concentration results in competition for CPU computational resources and network bandwidth, ultimately reducing the overall efficiency of the checkpointing process. To address this issue, PCache implements a strategy to distribute DP group checkpoint writing across different physical nodes, rather than concentrating them on a subset of nodes. By dispersing the write nodes for each DP group, this approach mitigates competition for computational resources and network bandwidth. In real-world experiments with a 5,000-accelerator MoE training task, this optimization reduced checkpoint writing latency by 50\%, while also lowering peak memory consumption on training nodes by 60\%, as is shown in Table~\ref{tab:checkpoint_cost}. 

\begin{table}[h]
\centering
\caption{Comparison of checkpoint save time costs (seconds).}
\vspace{3mm}
\begin{tabular}{l!{\vline}c!{\vline}c}
\toprule
\textbf{Test Case} & \textbf{\shortstack{PCache \\ (cost time)}} & \textbf{\shortstack{GPFS \\ (cost time)}} \\
\hline
\makecell{Megatron \\(tp=1 ep=8 pp=1, 128 accelerators)} & 70s & 160s \\ 
\midrule
\makecell{Megatron \\(tp=2 ep=8 pp=8, 512 accelerators)} & 90s & 240s \\
\bottomrule
\end{tabular}
\label{tab:checkpoint_cost}
\end{table}
\subsubsection{Cross-Cluster Synchronization Mechanism.}
In distributed AI training scenarios, achieving efficient and reliable data synchronization across cluster environments presents unique challenges. To address this, we developed Babel, a data synchronization middleware specifically designed for large-scale model training. Babel is tailored to solve the complex problem of efficiently transmitting massive unstructured datasets and high-frequency checkpoint (ckpt) files in cross-cluster and cross-region environments.

Whether it involves petabyte-scale datasets, billions of files, or frequently updated training state files, Babel provides a stable, high-speed, and accurate data synchronization service. Leveraging innovative features such as an adaptive data sharding strategy, efficient metadata prefetching mechanisms, and multi-dimensional data verification techniques, Babel significantly improves the transmission efficiency of large files while ensuring end-to-end data consistency. These capabilities provide robust technical support for distributed training environments. The main capabilities of Babel are summarized as follows.

\begin{figure}[h]
\begin{center}
\vspace{3mm}
\includegraphics[width=0.6\textwidth]{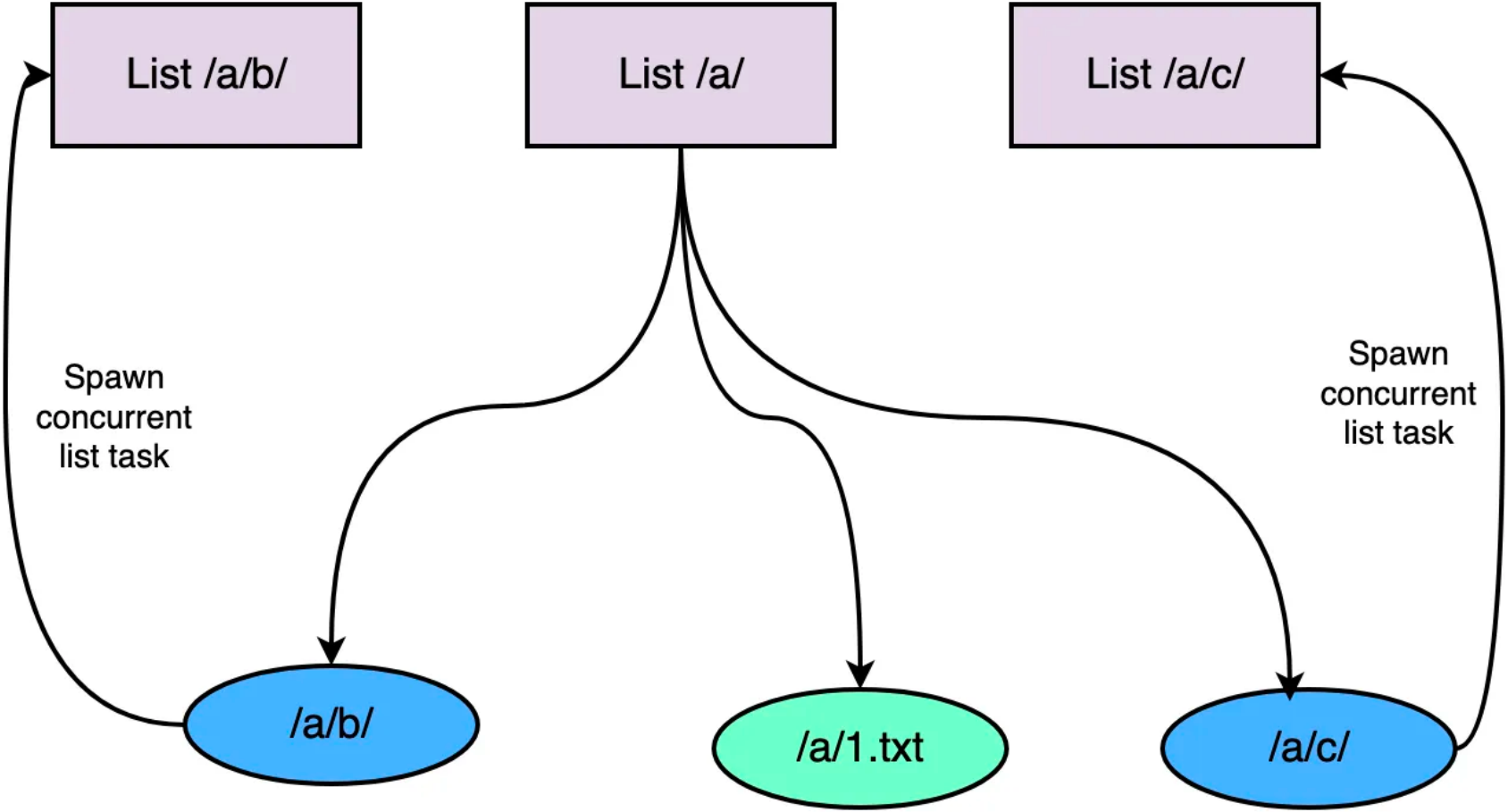}
\caption{The parallel metadata prefetching mechanism.}
\label{parallel_metadata_prefetching_mechanism}
\end{center}
\end{figure}

\paragraph{Metadata Prefetching Mechanism.}
In distributed training workflows, metadata management plays a pivotal role in determining task startup time. Synchronizing millions or even billions of file entries using traditional serial loading methods often leads to prolonged startup times, thereby significantly reducing overall efficiency. To address this challenge, we propose a parallel metadata prefetching mechanism (see Figure \ref{parallel_metadata_prefetching_mechanism}), which leverages concurrent Object Storage Service (OSS) List operations in conjunction with intelligent scheduling algorithms. 

To evaluate the effectiveness of this method, we conducted a performance test using OSS data comprising 190 million files. The results demonstrate a substantial improvement, with approximately a 36-fold increase in performance. Specifically, serial file listing required over six hours, whereas the parallel metadata prefetching mechanism reduced the processing time to approximately ten minutes. These findings highlight the significant efficiency gains achieved through the adoption of concurrent operations in large-scale distributed training environments.

\paragraph{Data Verification Technology.}
Babel implements a highly efficient and robust data verification framework that encompasses both metadata validation and content-based sampling cyclic redundancy check (CRC) verification. The system offers two distinct verification modes: real-time (runtime) verification and post-transfer verification. Traditional methods, such as MD5 hashing, often require extensive computational resources and significant time to verify large files (e.g., 100GB files), taking tens to hundreds of seconds to complete. To address these inefficiencies, Babel leverages a content-sampling-based CRC verification approach specifically designed for large files. This method significantly accelerates the verification process while reducing CPU consumption, all without compromising transmission accuracy. By adopting this optimization strategy, Babel reduces the verification time for a 100GB file to approximately three seconds, achieving an effective balance between verification speed and reliability.

\subsection{High-Efficiency Offline Inference Framework}
The current mainstream inference frameworks are primarily designed for online inference, emphasizing increased throughput under specific latency constraints. Consequently, parallelization strategies predominantly include inter-node tensor parallelism (TP) and intra-node pipeline parallelism (PP) across multiple computational nodes. However, TP often incurs substantial communication overhead, particularly in computing systems that lack high-speed interconnects such as NVLINK. In such cases, communication overhead can account for more than half of the total execution time. To address these limitations and enhance throughput by reducing communication overhead, we propose an efficient offline inference framework named Flood \cite{flood2025github}, which adopts a fully pipeline-parallel (PP) architecture.

Under high-concurrency scenarios, PP demonstrates superior throughput compared to TP and simplifies model adaptation by eliminating the need for tensor splitting. Instead of the conventional one-to-one mapping of processes to accelerators, our framework employs a many-to-one mapping strategy. This approach reduces inter-process communication overhead while offering greater flexibility in system design. To further isolate the performance impact of multiple processes on a single accelerator, a multi-stream strategy is implemented, where each process running on an accelerator is associated with a distinct stream. For multi-node inference, we initialize a number of processes on each node equal to the total number of pipeline stages. Leveraging the PP strategy enables the achievement of zero CPU overhead. For instance, in a single-node configuration with 8 accelerators, we deploy 9 processes such that there is always one process waiting for the accelerator assigned to the first pipeline stage to become available. This ensures that accelerator resources are utilized continuously, thereby minimizing idle time.

\begin{figure}[ht]
	\centering
	\includegraphics[width=0.85\textwidth]{"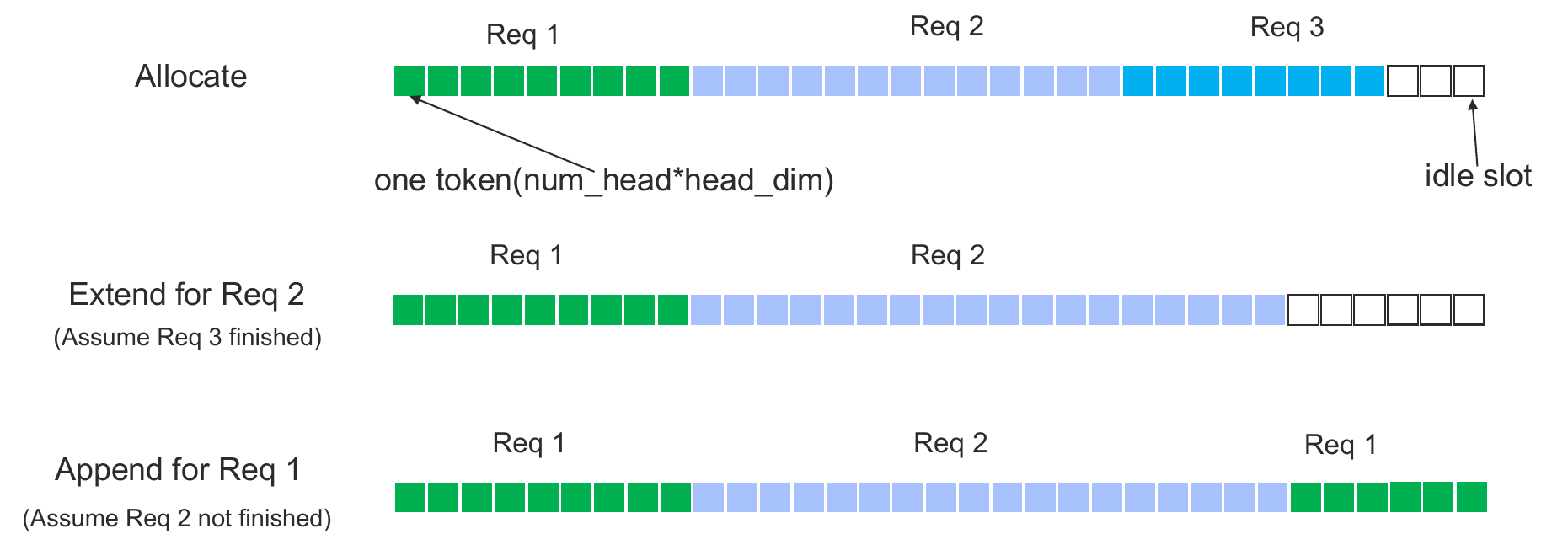"}
	\caption{Segment kvcache.}
	\label{fig:segment_kvcache}
\end{figure}

In parallel, popular frameworks such as vLLM \cite{kwon2023efficientmemorymanagementlarge} commonly utilize block tables for managing the key-value cache (kvcache). However, small block sizes can result in inefficient utilization of computational resources. To address this issue and maximize accelerator resource usage, we propose a novel segment cache mechanism that allocates the kvcache in a contiguous memory space to enable the use of larger block sizes. Specifically, we allocate a kvcache tensor with the shape $[max\_token\_num, num\_head, head\_dim]$. During inference, a pre-allocated contiguous memory space is dedicated to each request to accommodate both the prompt and the output. This design, illustrated in Figure \ref{fig:segment_kvcache}, facilitates efficient memory management and enhances computational performance.

In typical scenarios where the user-defined maximum output length is relatively small, the cache can be allocated based on this predefined maximum length. However, in cases where the specified maximum output length is exceptionally large (e.g., 32,768 tokens) and significantly surpasses the actual generated output length, this can result in the allocation of overly large segment caches, thereby reducing request concurrency. To address this inefficiency, a segment with a conservative size will be used during the initial allocation phase. If the actual generated tokens exceeds the allocated segment size, the following strategies can be employed:

\begin{itemize}
\item \textbf{Extend the current segment.} If the next segment in the kvcache memory space is free, the current segment can be extended into the adjacent space.
\item \textbf{Append an additional segment.} If the next segment is occupied and an other segment is available, it can be appended to the request’s segment list to accommodate the overflow.
\item \textbf{Wait.} If neither extension nor appending is possible, the request is placed in a wait-list until a segment becomes available.
\end{itemize}

The segment cache not only resolves the challenges associated with excessively long maximum output lengths but also inherently supports prefix caching. For batch requests sharing a common prefix, the prefix can be stored using a single segment or a combination of multiple segments.

Finally, we compare our Flood with vLLM (the version is `0.6.6.post2') on a benchmark dataset, i.e., shareGPT \cite{sharegpt2023datasethuggingface}, and the detailed performance comparison is listed in Table \ref{tab:model_performance}. The performance is measured by generated tokens per second.

\begin{table}[]
    \centering
    \caption{Inference performance comparison (the device details are listed in Table \ref{tab:computational_resources}).}
    \vspace{3mm}
    \tabcolsep=0.2cm
    \resizebox{0.75\linewidth}{!}{%
        \begin{tabular}{l|c|c|c|c}
        \toprule
        \textbf{Model} & \textbf{Device} & \textbf{vLLM (token/s)} & \textbf{Flood (token/s)} & \textbf{Speedup} \\
        \midrule
        Ling-Lite &  1 * Device E &  4355 & 5869 & 1.35 \\
        % Ling-Lite &  2*L20 &  3523 & 6140 & 1.74  \\
        Ling-Lite &  1 * Device C & 3576 & 5451 & 1.52 \\
        \midrule
        Ling-Plus & 16 * Device B & 2331 & 4857 & 2.08 \\
        Ling-Plus(FP8) & 8 * Device E &  2742 & 6569 & 2.40 \\
        \bottomrule
        \end{tabular}
    }
    \label{tab:model_performance}
\end{table}
	\section{Pre-Training}\label{section_pre_training}

\subsection{Pre-Training Data}
The Ling models demonstrate their competitive performance through rigorous methodologies designed to enhance the quality of large-scale pre-training datasets. The corpus utilized in the model development is a diverse collection of textual and non-textual data, encompassing sources such as web content, books, academic papers, social media, encyclopedias, mathematics, and programming code. To date, we have constructed a high-quality corpus consisting of approximately 9 trillion tokens, distributed across 1 trillion tokens in Chinese, 5.5 trillion in English, and 2.5 trillion in code. The development of such a large-scale, high-quality dataset is the result of systematic improvements in several key areas:

\begin{itemize}
    \item \textbf{Data curation.} The majority of raw data used in this study were obtained from publicly available sources, including Common Crawl (CC), coding platforms, and encyclopedias. However, these sources often exhibit a range of quality issues. To address this, we developed specialized data cleaning pipelines tailored to the characteristics of different data types (e.g., web pages, academic papers, books, and code). The cleaning process included tasks such as text extraction and parsing from raw HTML/PDF files, deduplication, rule-based filtering, and the removal of toxic or undesirable content. Furthermore, we established a robust quality assessment framework comprising 10 categories and over 300 quality evaluation metrics. This framework enables us to systematically categorize datasets into quality tiers, which serve as a foundation for further refinement and the informed selection of training data.
    \item \textbf{High-quality data selection.} To identify high-quality data, we fine-tuned models such as fastText \cite{bojanowski2017enrichingwordvectorssubword} and BERT \cite{devlin2019bertpretrainingdeepbidirectional}, applying fine-grained labels and attributes to the cleaned data. These attributes include metrics such as text coherence, knowledge density, educational level, and complexity. Using this approach, we were able to extract high-quality data samples from the broader dataset. Additionally, sampling experiments were conducted across various features to identify optimal strategies for data selection. This process ensured that the selected data were well-suited for enhancing downstream model performance.
    \item \textbf{Mathematics and code data.} Informed by prior research \cite{shao2024deepseekmathpushinglimitsmathematical}, we curated a large-scale dataset focused on programming and mathematical reasoning. The data collection process leveraged publicly available repositories such as Common Crawl and GitHub. To ensure the quality and relevance of the mathematics and code data, we developed advanced filtering models based on fastText and BERT. These models were employed to identify and retrieve content containing high-quality programming and mathematical reasoning, enabling the incorporation of specialized knowledge into the corpus.
    \item \textbf{Data ablation and mixture.} Building on insights from earlier studies \cite{deepseekai2024deepseekcoderv2breakingbarrierclosedsource}, we employed a continued-training strategy to assess the contribution of newly integrated datasets to the overall model performance. This process was conducted on smaller models to validate the utility of the new data prior to full-scale implementation. Additionally, our data mixing strategy prioritized diversity and ensured balanced distributions across different data attributes. Particular attention was given to optimizing sampling strategies for critical data types, such as reasoning-related content. The effectiveness of these strategies was further validated on larger models, demonstrating their impact on enhancing training outcomes.
\end{itemize}

\subsection{Model Architecture}
In contrast to conventional dense architectures, the MoE paradigm replaces standard FeedForward Networks (FFNs) with a collection of $N$ experts \cite{fedus2022switch, lepikhin2020gshard, jiang2024mixtral}, which are compact and modular FFN units. This design enables greater efficiency (shown in the following subsection \ref{sec:scaling}) and specialization within LLMs.
The core mechanism of MoE is the dynamic routing of tokens to specific experts through an individual router, $\mathrm{R}$, for each token. This routing facilitates highly optimized and selective computation, as defined by the following equations:

\begin{equation}
\begin{split}
    &\mathbf{p}_t = \mathrm{Softmax}(\mathrm{R}(\mathbf{h}_t)),\\
    &\mathbf{o}_t = \sum_i \mathbf{p}_{t,i}\mathrm{E}_i(\mathbf{h}_t) \quad\text{s.t.} \quad \mathbf{p}_{t,i} \in \mathrm{Topk}(\mathbf{p}_t).
\end{split}
\end{equation}
where $\mathbf{h}_t \in \mathbb{R}^d$ is the $d$-dimensional FFNs input of the $t$-th token, $\mathrm{E}_i$ is the $i$-th expert in total $N$ experts, $\mathbf{p} \in \mathbb{R}^N$ denote the gates for expert selection, and $\mathbf{o}_t \in \mathbb{R}^d$ denotes the output of the $t$-th token after being processed by routing experts.
Next, we will elaborate Ling's architectural innovations on the above MoE framework.

\subsubsection{Fine-Grained Experts}
To enhance the advantages of the MoE architecture over traditional dense models, while simultaneously improving training efficiency and scalability, the Ling models adopt a fine-grained expert strategy \cite{dai2024deepseekmoe, deepseekai2024deepseekv2strongeconomicalefficient}.
Specifically, compared to the original expert design, our approach scales the number of experts while proportionally reducing the intermediate size of each expert, thus maintaining the equivalent total capacity.
This design promotes a higher degree of specialization among experts, allowing the model to encapsulate a wider and more diverse range of knowledge.

Nevertheless, solely relying on fine-grained experts poses a potential challenge, 
i.e., individual experts may struggle to simultaneously develop both general and specialized capabilities under constrained capacity.
This limitation may incentivize experts to prioritize improving general capabilities over specialized ones, which contradicts the design intent of the fine-grained experts.
To address this, we introduce an additional share expert that can utilize all tokens for training without the need of routing \cite{rajbhandari2022deepspeed, dai2024deepseekmoe} to provide general ability. The final output $\mathbf{o}_t'$ of the MoE FFNs can be represented as follows:

\begin{equation}
    \mathbf{o}_t' = \mathbf{o}_t + \mathrm{E_{share}}(\mathbf{h}_t).
\end{equation}

\subsubsection{Expert Routing}
Routing in MoE-based LLMs can generally be categorized into two approaches: token-drop and dropless. To ensure the efficient utilization of training data, we adopt a dropless strategy in our implementation. Additionally, we incorporate two key mechanisms, i.e., load balance loss and router z-loss, to enhance training efficiency and to prevent imbalances in the distribution of tokens across experts.

Meanwhile, to mitigate the instability issue during the early stage of pretraining, we propose a \emph{Stochastic Routing Warmup} mechanism.
Unlike conventional load-balancing losses or manual interventions, this method introduces controlled randomness into the routing module to prevent expert overload,
and further prevent the experts from collapsing due to routing imbalance during the early training stage.
Let $\mathbf{s}_t \in \mathbb{R}^N$ denote the raw routing logits for input token $t$, computed by a linear projection layer. During the warmup phase (global step $i \leq W$), we interpolate between learned logits and synthesized random logits. The final routing logits $\hat{s}_t$are computed as:

\begin{equation}
\begin{split}
    &\mathbf{\hat{s}}_t = \alpha \cdot \mathbf{s}_t + (1 - \alpha) \cdot (\mu_s + \sigma_s \cdot \epsilon),\\
    &\alpha = \min(\frac{i}{W}, 1.0),
    \quad \epsilon \sim \mathcal{N}(0, I),
\end{split}
\end{equation}

where $\mu_s$ and $\sigma_s$ represent the running mean and standard deviation of $\mathbf{s}_t$. 
The router warmup ensures balanced expert activation at initialization while gradually shifting control to the learned routing distribution, effectively mitigating out-of-memory risks and stabilizing training process.  

\subsubsection{NormHead}
Compared to dense architectures, MoE-based models exhibit increased training complexity and significantly reduced stability, which can lead to fluctuations in loss values and hinder convergence. 
During our preliminary experiments, we observed that the output norm of the LM-Head often becomes unstable, particularly during loss spikes. This instability can negatively impact both the convergence and the overall performance of the model.
To address this issue, we involve a Normed LM-Head (NormHead) for token prediction \cite{yang2023baichuan}. In this approach, the weight of the LM-Head, i.e., $\mathbf{W}_{lm\_head}$ are subjected to L2 normalization before being applied for the token prediction. The formulation is as follows:
\begin{equation}
    \mathbf{h}_o = \frac{\mathbf{W}_{lm\_head}}{||\mathbf{W}_{lm\_head}||_2} \mathbf{h},
\end{equation}
where $\mathbf{h}$ represents the input to the LM-Head, and $\mathbf{h}_o$ is the normalized output. By normalizing the weights, the NormHead ensures that variations in weight magnitude do not contribute to instability, particularly in scenarios involving large gradients or fluctuating loss values. Our empirical experiments indicate that NormHead significantly enhances the stability of training.

\subsection{Scaling Laws}
\label{sec:scaling}
As a foundational principle, \emph{scaling laws} provide valuable predictive insights into the behavior of LLMs as model capacity and data volume increase. These scaling laws can serve as a framework for comparing different LLM architectures and guiding the training of hundred-billion-parameter LLMs. The existing studies \cite{kaplan2020scaling,hoffmann2022training,henighan2020scaling} have extensively investigated scaling laws in the context of dense LLMs, whereas some subsequent works \cite{gao2024scaling, clark2022unified} have initiated discussions on the scaling laws for MoE models. However, there remains a significant gap in the systematic exploration of scaling laws for MoE architectures.

In developing the MoE architecture for the Ling models, we conducted a systematic analysis of its scaling behavior with respect to two critical hyper-parameters: batch size and learning rate. Additionally, we evaluated its overall performance by examining validation loss. Using the scaling law of loss as a framework, we compared the MoE and dense architectures, uncovering insights into the scaling behavior and effectiveness of MoE models.

\subsubsection{Scaling Laws for Hyper-parameters}
To optimize model performance across varying compute budgets, it is essential to first determine the optimal batch size and learning rate for different model sizes and data scales. To achieve this, we aligned the MoE architecture with Ling-Plus during scaling law experiments to minimize variability in the factors influencing model performance. We then performed a grid search over batch size and learning rate in the context of small-scale MoE model training, with compute budgets spanning a range from $1e^{18}$ to $6e^{20}$. This systematic exploration enabled us to identify configurations that maximize the efficiency and performance of the models under differing computational constraints. 

\begin{figure}[ht]
	\centering
	\begin{subfigure}{.5\textwidth}
		\includegraphics[width=1.0\textwidth]{"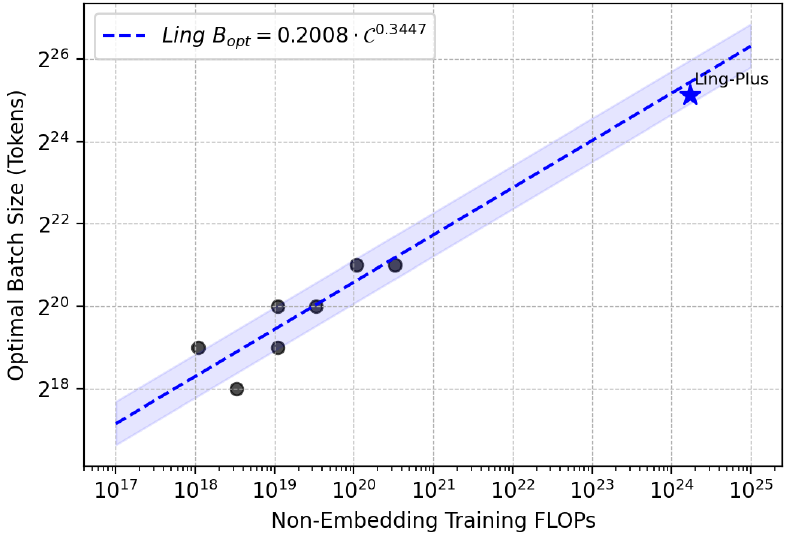"}
		\caption{Batch size.}
		\label{fig:bs}
		%\vspace{-0.5em} 
	\end{subfigure}%
	\begin{subfigure}{.5\textwidth}
		\centering
		\includegraphics[width=1.0\linewidth]{"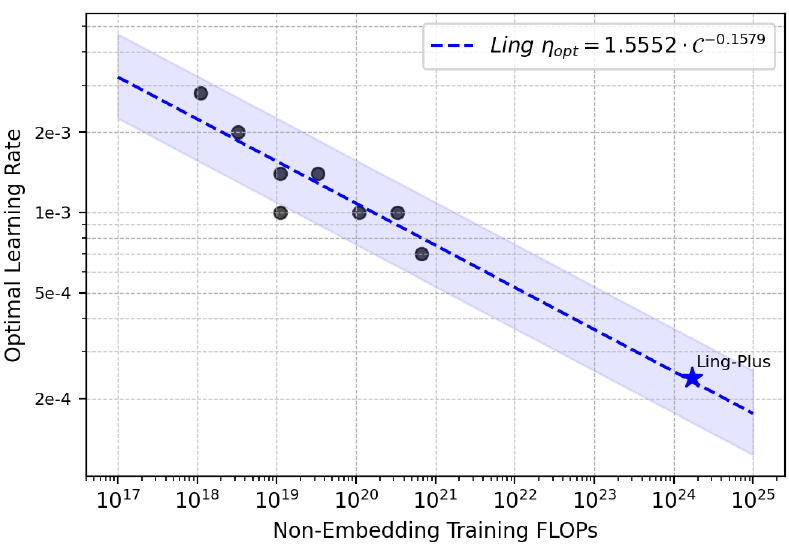"}
		\caption{Learning rate.}
		\label{fig:lr}
	\end{subfigure}%
	\caption{Scaling curve for batch size and learning rate.}
	\label{fig:bslr}
\end{figure}

Subsequently, we modeled the power law relationship between batch size ($B$) and learning rate ($\eta$) with respect to the compute budget ($\mathcal{C}$). The results are illustrated in Figure \ref{fig:bslr} and provide insights into how these hyper-parameters scale under varying computational constraints.

The fitted results indicate that the scaling behaviors of batch size ($B$) and learning rate ($\eta$) for MoE models are consistent with those observed in dense models, aligning with findings from previous work \cite{deepseekai2024deepseekllmscalingopensource}. Furthermore, we adjusted the MoE architecture, specifically the number of routed and shared experts, to achieve varying degrees of sparsity, ranging from 4.6\% to 12.1\%. In addition, we tuned the weighting of auxiliary loss components, such as balance loss and z-loss, to evaluate their potential impact.

Our analysis revealed that, for a given compute budget, neither the MoE architecture nor the auxiliary loss functions had any significant influence on the optimal batch size and learning rate. Instead, the optimal configuration of these two hyper-parameters was found to be primarily determined by the compute budget. This observation underscores the compute budget as the dominant factor in hyper-parameter tuning for MoE training.

\subsubsection{Scaling Laws for Model Performance}
One of the critical questions surrounding the MoE architecture is its efficiency relative to dense architectures. To address this, we define the \emph{efficiency lever} of MoE compared to dense models as the ratio of compute budgets required for each architecture to train a sufficiently large model that achieves the same level of training loss. To evaluate this, we selected a range of compute budgets spanning from $1e^{18}$ to $3e^{20}$ and conducted small-scale training experiments for both MoE and dense models. For each compute budget, we collected the results corresponding to the optimal training loss and subsequently fit a logarithmic inverse FLOPs-to-Loss curve for both MoE and dense architectures, respectively. This approach enabled a systematic comparison of the training efficiency between the two architectures.

\begin{figure}[ht]
	\centering
	\includegraphics[width=0.6\textwidth]{"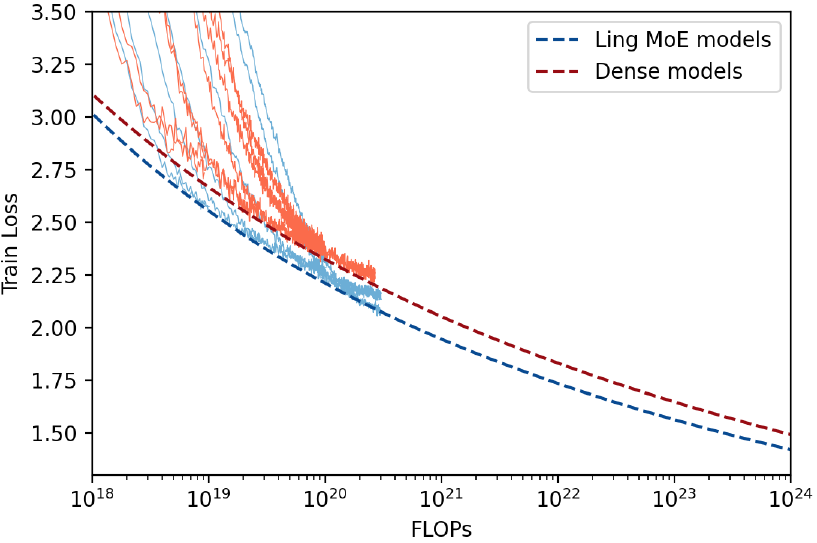"}
	\caption{Scaling curve of loss.}
	\label{fig:loss}
\end{figure}

As evidenced in Figure \ref{fig:loss}, MoE architecture consistently achieves lower training loss than the dense architecture under equivalent compute budgets.
The average efficiency lever is approximately 3x, meaning that the MoE architecture is about 3 times more efficient than the dense architecture in terms of compute required to achieve the same performance. Interestingly, we observed that the efficiency lever increases as the compute budget grows. For example, at $1e^{21}$ FLOPs, the efficiency lever is approximately 3, whereas at $1e^{24}$ FLOPs, the efficiency lever exceeds 3.5. 
This trend suggests that MoE architectures exhibit increasing advantages over dense architectures as the compute budget scales up, reinforcing their potential for large-scale applications.
The enhanced scalability implies that MoE architectures, such as Ling, could become even more powerful and efficient when applied to massive-scale models, providing significant benefits in terms of resource utilization and performance at higher compute budgets. As models continue to scale, these efficiency gains highlight the promise of MoE as a highly scalable and cost-effective alternative to dense architectures.

\subsection{Training Recipe}

\subsubsection{Initial Pre-Training}

We pre-train Ling-Plus using the AdamW optimizer with the following hyper-parameters: $\beta_1 = 0.9$, $\beta_2 = 0.95$, $\epsilon = 1e^{-8}$, and \verb|weight_decay| = 0.1. We adopt a warmup-and-stable-decay learning rate schedule, with a maximum learning rate of $2.4e^{-4}$. The learning rate is linearly warmed up from 0 to the maximum value over the first 2K training steps. Afterward, it is halved once approximately 60\% of the training tokens are processed.

We also implement a batch size warmup strategy, starting from an initial batch size of 2,560. The batch size gradually increases to a maximum of 8,960 and remains at this maximum for the remainder of training. The gradient clipping norm is set to 1.0, and the maximum sequence length is fixed at 4K tokens. For the first stage of pre-training, we train on a total of 9T tokens. The load-balancing loss coefficient is set to $0.015$, and the z-loss coefficient is set to $1e^{-4}$. We do not employ the token-dropping strategy during training.

Throughout the training process, we continuously monitor various indicators such as training loss, gradients, router token distribution, and benchmark scores to ensure that the model is learning effectively and consistently. We also perform several adjustments to the pre-training data mix to enhance model performance. During each adjustment, we increase the proportion of high-quality data while removing samples where the loss fails to decrease. To mitigate the risk of duplicate samples during these adjustments, we employ sample-level online data deduplication, ensuring the uniqueness of training data during the mixing process. These techniques and strategies collectively aim to optimize the pre-training process, ensuring robust model performance while maintaining training stability and data quality.

\subsubsection{Long Context Pre-Training}
During this phase of pre-training, the maximum input sequence length was extended to 16K tokens. This extension was achieved using Rotary Position Embedding (RoPE), with $\theta$ parameter adjusted from 10K to 600K to support longer sequences. Adjustments were also made to the training dataset to better align with the objectives of long-context processing. For the Ling-Plus model, the proportion of web-derived data was reduced, and additional long-form text data were incorporated to improve the model's ability to process extended sequences. Similarly, for the Ling-Lite model, the amount of web-based data was scaled down, while the proportion of mathematical and coding-related corpora was increased. The learning rate schedule remained consistent with the prior training stage, and a total of 150B tokens were processed during this phase to enhance the model's long-context processing capabilities.

\subsubsection{Annealing}
During the final phase of pre-training, the inverse square root decay schedule was employed to systematically reduce the learning rate from $1.2e^{-4}$ to $1.2e^{-8}$. To maintain the effectiveness of this phase, the annealing process was conducted exclusively using clean, meticulously curated, high-quality datasets.

\subsubsection{Skip loss spikes and Sample retry mechanism}\label{sec:loss_spikes}
\begin{figure}[h]
    \centering
    \begin{subfigure}{.49\textwidth}
            \centering
    	\includegraphics[width=1.0\textwidth]{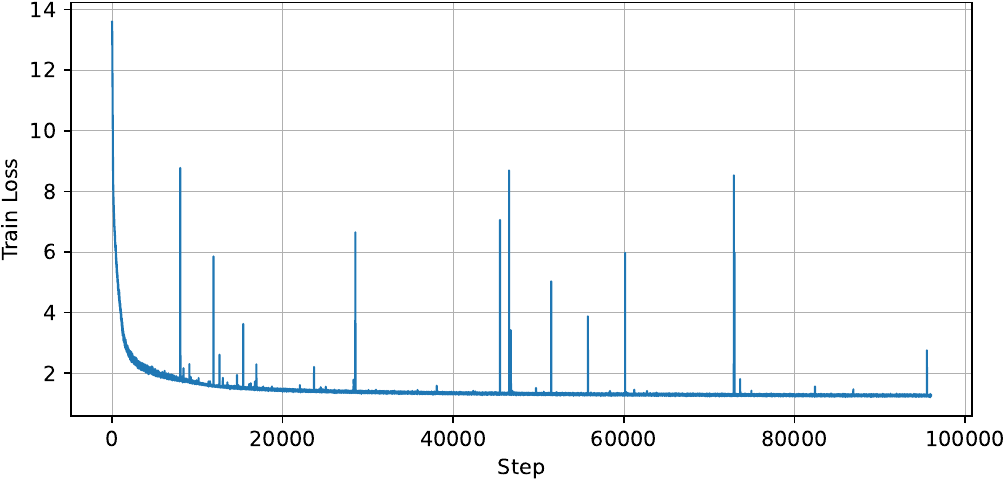}
    	\caption{before}
            % \caption{}
            \label{fig:llm_spike_skip_and_retry_raw_loss}
    \end{subfigure}
    \begin{subfigure}{.49\textwidth}
            \centering
    	\includegraphics[width=1.0\textwidth]{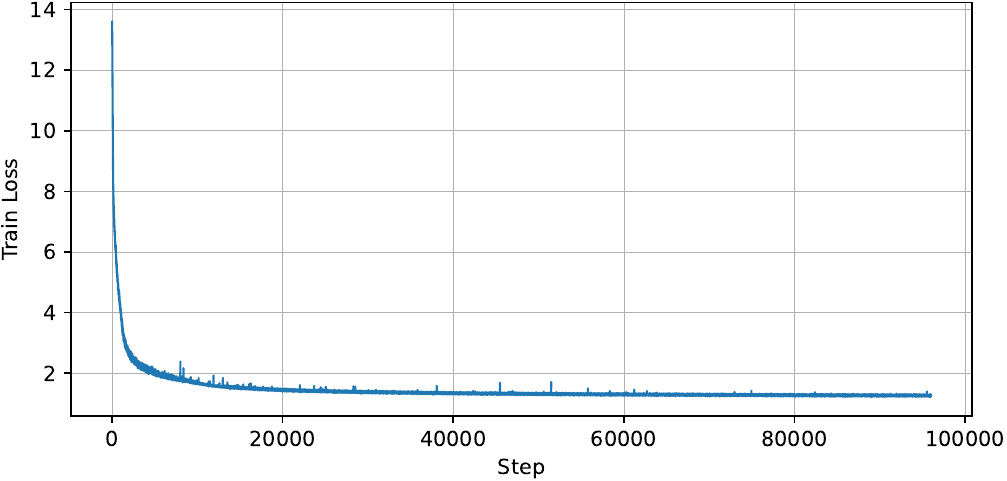}
    	\caption{after}
            % \caption{}
            \label{fig:llm_spike_skip_and_retry_new_loss}
    \end{subfigure}
    \caption{Comparison of train loss curves before and after applying skip loss spikes and sample retry mechanism.}
    \label{fig:llm_spike_skip_and_retry}
\end{figure}

During pre-training, the phenomenon where the loss abruptly rises and then falls is referred to as loss spikes.
These abrupt changes are typically triggered by specific interactions between the data and optimizer states. Loss spikes can be classified into two types: (1) \textit{narrow spikes}, which last for only a few steps and have minimal impact on model performance, and (2) \textit{wide spikes}, which persist across more steps and can significantly disrupt model stability, sometimes even causing benchmark evaluation results to approach random levels. Our research shows that it is difficult to completely eliminate loss spikes. To mitigate their effects, we have designed a series of strategies, including skip loss spikes and sample retry mechanism. When a loss spike is detected, the affected update is skipped, and the associated data is randomly re-injected into subsequent training batches. If the spike persists, we automatically reduce the learning rate during the affected step. This approach has proven effective in reducing the negative impact of loss spikes, enabling consistent improvements in benchmark metrics throughout the training process.
Figure \ref{fig:llm_spike_skip_and_retry} intuitively illustrates the improvement in train loss achieved by the proposed strategy.
	\begin{figure}[htp]
\begin{center}	
\includegraphics[width=0.9\textwidth]{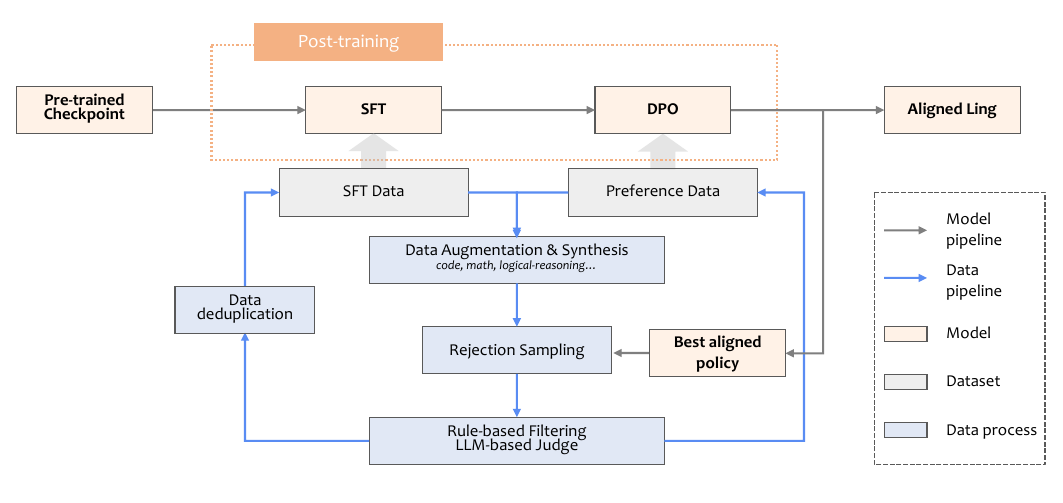}
\caption{Illustration of our post-training modeling and dataset curation pipeline.}
\label{fig:post_training}
\end{center}
\end{figure}
\section{Post-Training}\label{section_post_training}
As depicted in Figure \ref{fig:post_training}, following the pre-training phase, the development of the Ling model involves a dual-stage alignment process: Supervised Fine-Tuning (SFT) \cite{ouyang2022training} and Direct Preference Optimization (DPO) \cite{RafailovSMMEF23_dpo}. This alignment process adopts an iterative framework to progressively enhance both the dataset and the model's capabilities. Specifically, after each training cycle, the best-performing model from prior iterations is leveraged to refine the SFT and preference data, thereby informing and improving subsequent training phases. In the following sections, we provide a detailed overview of our SFT data curation process (see Section \ref{sft}), the DPO methodology (see Section \ref{dpo}), as well as the post-training techniques employed for long-text generation (see Section \ref{long}) and tool use (see Section \ref{tool}). Additionally, the complete post-training pipelines and configurations, tailored for various computing platforms, will be made publicly available through our code repository.

\subsection{Supervised Fine-tuning}
\label{sft}
Data serves as the cornerstone of the supervised fine-tuning (SFT) stage, with synthetic data assuming an increasingly significant role. This shift is driven by the diminishing availability of human-generated data and the substantial costs associated with its annotation, both in terms of time and labor. Our SFT dataset is constructed from an initial seed dataset consisting of one to two million instances derived from a combination of human annotations and open-source resources. This dataset is subsequently scaled by a substantial factor through the application of data synthesis techniques. In the following, we detail the measures implemented to ensure the quality and diversity of the synthetic data, which are critical to the success of the fine-tuning process.

\subsubsection{Quality Assurance}
Our data synthesis process leverages methodologies inspired by Magpie-like approaches \cite{magpie} and OSS-Instruct \cite{wei2023magicoder} to generate novel problem prompts. Following this, we applied the rejection sampling (RS) technique, as outlined in \cite{llama3}, to produce candidate responses. To ensure high-quality outputs, we established a dedicated pipeline designed to select the most optimal responses, with specific tailoring for reasoning and non-reasoning datasets. This targeted approach ensures that the synthesized data aligns with the desired quality standards while addressing the diverse requirements of different data types.

For \emph{reasoning} data, such as code, math, and logical reasoning, we implemented a series of rule-based filtering mechanisms to ensure high-quality data:
\begin{itemize}
    \item \textbf{Code data.} A comprehensive multi-stage validation process was developed, which involves: (1) extracting and verifying code through rule-based checks and execution tests, (2) generating synthetic test cases using advanced LLMs, and (3) retaining only those code solutions that successfully pass all stages of verification.
    \item \textbf{Math data.} We prompted LLMs to translate computational logic into executable Python code, enabling the execution of the code to verify both the final answers and intermediate reasoning steps. This ensures accuracy throughout the problem-solving process.
    \item \textbf{Logical reasoning data.} A majority voting system was employed to achieve consensus-based selection of the best solutions. This approach helps identify the most accurate logical conclusions by aggregating multiple perspectives to ensure reliability.
\end{itemize}
Our empirical analysis indicates that these rule-based methods effectively eliminated a significant number of incorrect responses while minimizing the exclusion of correct ones, thus maintaining dataset integrity. Subsequently, for both the filtered reasoning data and \emph{non-reasoning} data (e.g., creative writing and general question answering), we employed an LLM-based judge with a detailed evaluation checklist to further assess the relevance and quality of the generated responses. This final quality control step ensures the overall robustness of the dataset.

\subsubsection{Data Redundancy} 
During the data synthesis process, we observed a degree of redundancy within the dataset, particularly in code-related synthetic data \cite{tsai2024code}. This redundancy often stemmed from similar response patterns, which posed the risk of causing the model to overfit to specific patterns, thereby potentially impairing its generalization capabilities. To address this issue, we implemented a semantic-based deduplication method to eliminate redundant data.

Specifically, we employed a text embedding model with demonstrated strong performance, as reported on the MTEB Leaderboard \cite{mteb2024leaderboard}, to map problem prompts and responses into a vector space. Using these embeddings, we identified and removed instruction data with high cosine similarity, effectively reducing redundancy in the dataset. Our analysis revealed that removing approximately 10\% to 20\% of the most similar data had no adverse effect on the model's core capabilities. This finding underscores the significant potential for deduplication within synthetic datasets to enhance data quality without compromising model performance.

\subsection{Direct Preference Optimization}\label{dpo}
The Direct Preference Optimization (DPO) workflow consists of two primary phases, each involving multiple iterative processes aimed at enhancing preference alignment and improving the robustness of the model's reasoning:
\begin{itemize}
        \item \textbf{Vanilla DPO (VD).} In this initial phase, the focus is on improving the model's authenticity, relevance, harmlessness, and capacity to follow instructions. Preference data is curated through a rejection sampling strategy that integrates scoring mechanisms from both a large language model (LLM) judge and a reward model.
        \item \textbf{Robustness optimization (RO).} The second phase emphasizes strengthening the stability of the model's reasoning across tasks such as mathematics, coding, and overall performance. For tasks with definitive answers, rejection sampling is utilized within a probability range of correct responses (e.g., 0.2 to 0.6), with responses selected via a Best-of-N approach and rejected using a Worst-of-N strategy based on reward model scores. For open-ended tasks, quality evaluation is performed through majority voting to mitigate bias and improve overall robustness. Additionally, a negative log-likelihood (NLL) regularization term~\cite{PangYHCSW24_nll_loss} with a weight of 0.05 is introduced. This regularization is designed to prevent high-quality selected responses from experiencing a decline in their probabilities, thereby maintaining output quality.
    \end{itemize}
    
To enhance training efficiency, we implemented an innovative data-packing strategy within the DPO framework. This method involves padding both chosen and rejected sequences to the maximum sequence length to maintain the integrity of the chosen-rejected pairing paradigm. By adopting this approach, we achieved a significant \textbf{3.7-fold increase in DPO training speed}. During the iterative optimization process, we identified issues related to the clarity and structural organization of responses, particularly in adherence to formatting instructions. To address these shortcomings, an additional DPO training phase focused specifically on formatting was conducted. This involved utilizing pairs of accepted and rejected responses that shared identical reasoning but differed in formatting. During the computation of the DPO loss, masking was applied to all content except the format-specific portions to ensure that valid reasoning within the rejected responses was not penalized. This precaution mitigated the risk of suppressing useful reasoning due to the contrastive nature of the DPO loss.

Empirical results, as presented in Table~\ref{tab:results_Ling_moe_plus_format_DPO}, demonstrate that format-focused training effectively reduces penalties stemming from formatting errors. This improvement enables the model's capabilities to be evaluated and utilized more reliably.

\begin{table}
    \centering
        \caption{The model's performance across various post-training stages on multiple benchmarks, considering specific formatting requirements. `DPO-format' refers to a DPO training designed for format recovery.}
    \vspace{3mm}
    \tabcolsep=0.2cm
    \resizebox{0.9\linewidth}{!}{%
        \begin{tabular}{l|c|c|c|c|c|c|c}
        \toprule 
        \textbf{Model}  & \textbf{AGIEval} & \multicolumn{1}{c|}{\textbf{CMATH}} & \multicolumn{1}{c|}{\textbf{MATH}} & \begin{tabular}[c]{@{}c@{}}\textbf{CN Middle}\\\textbf{School 24}\end{tabular} & \textbf{GaoKao} & \begin{tabular}[c]{@{}c@{}}\textbf{Olympiad}\\\textbf{Bench}\end{tabular} & \begin{tabular}[c]{@{}c@{}}\textbf{Minerva}\\\textbf{Math}\end{tabular} \\
        \midrule
        SFT & 64.78 & 94.72 & 78.98 & 59.41 & 51.65 & 44.30 & 40.81 \\
        DPO & 65.76 & 95.26 & 80.16 & 70.30 & 57.14 & 43.41 & 40.07 \\
        DPO-format & \textbf{67.67} & \textbf{95.63} & \textbf{80.62} & \textbf{73.27} & \textbf{63.74} & \textbf{44.89} & \textbf{41.18} \\
        \bottomrule
        \end{tabular}
    }
        \label{tab:results_Ling_moe_plus_format_DPO}
\end{table}
    
In the early stages of model development, we conducted experiments with various approaches to improve performance. Below, we present key insights from these exploratory attempts to inform future research directions:

    \begin{itemize}
        \item \textbf{Length-regularized DPO.} We explored incorporating length regularization into the DPO framework to mitigate the model's sensitivity to response length. While this approach effectively shortened response outputs, it did not yield improvements in overall performance. In fact, the vanilla DPO method outperformed length-regularized DPO, particularly on tasks involving mathematics and coding. We hypothesize that this limitation arises because complex problems in these domains often require detailed and lengthier solutions. Length regularization may inadvertently suppress the loss for such responses, reducing the model's ability to handle intricate cases that inherently demand more extensive outputs. These findings highlight a trade-off: although length regularization can help control verbosity, it risks penalizing the longer responses necessary for addressing complex tasks. Future research could investigate adaptive strategies that balance length control with the varying complexity and requirements of tasks across different domains.
        \item \textbf{Avoid repeated prompts.} We also experimented with augmenting the training dataset by incorporating repeated prompts paired with varied responses. The goal was to expand the model's exploration space and potentially enhance downstream DPO performance. This approach involved generating additional responses by sampling at different temperature settings. However, our results showed a slight performance decline compared to using the original dataset. This suggests that increasing data volume through repeated prompts does not necessarily lead to better optimization outcomes in DPO. Instead, our findings indicate that prioritizing diverse and unique prompts is more effective for improving performance.
        
    \end{itemize}

\subsection{Long Context}
\label{long}
The Ling model is designed to process text lengths of up to 16k tokens, addressing the requirements of long-form content processing. To enhance the model's performance on long-context tasks while ensuring that its capabilities on shorter tasks remain unaffected, it is essential to carefully refine the training strategy during the post-training phase. To achieve this goal, the following efforts were undertaken to strengthen the model's ability to handle extended contexts:

\subsubsection{Synthesis of Long-Context Instruction Data}
We curated high-quality Chinese and English documents from open-source corpora and concatenated them to create extended contexts. Using these extended contexts, the model was prompted to generate queries and responses across a variety of tasks, including retrieval, summarization, question answering (QA), and reasoning.

\begin{figure}[ht]
	\centering
	\includegraphics[width=0.8\textwidth]{"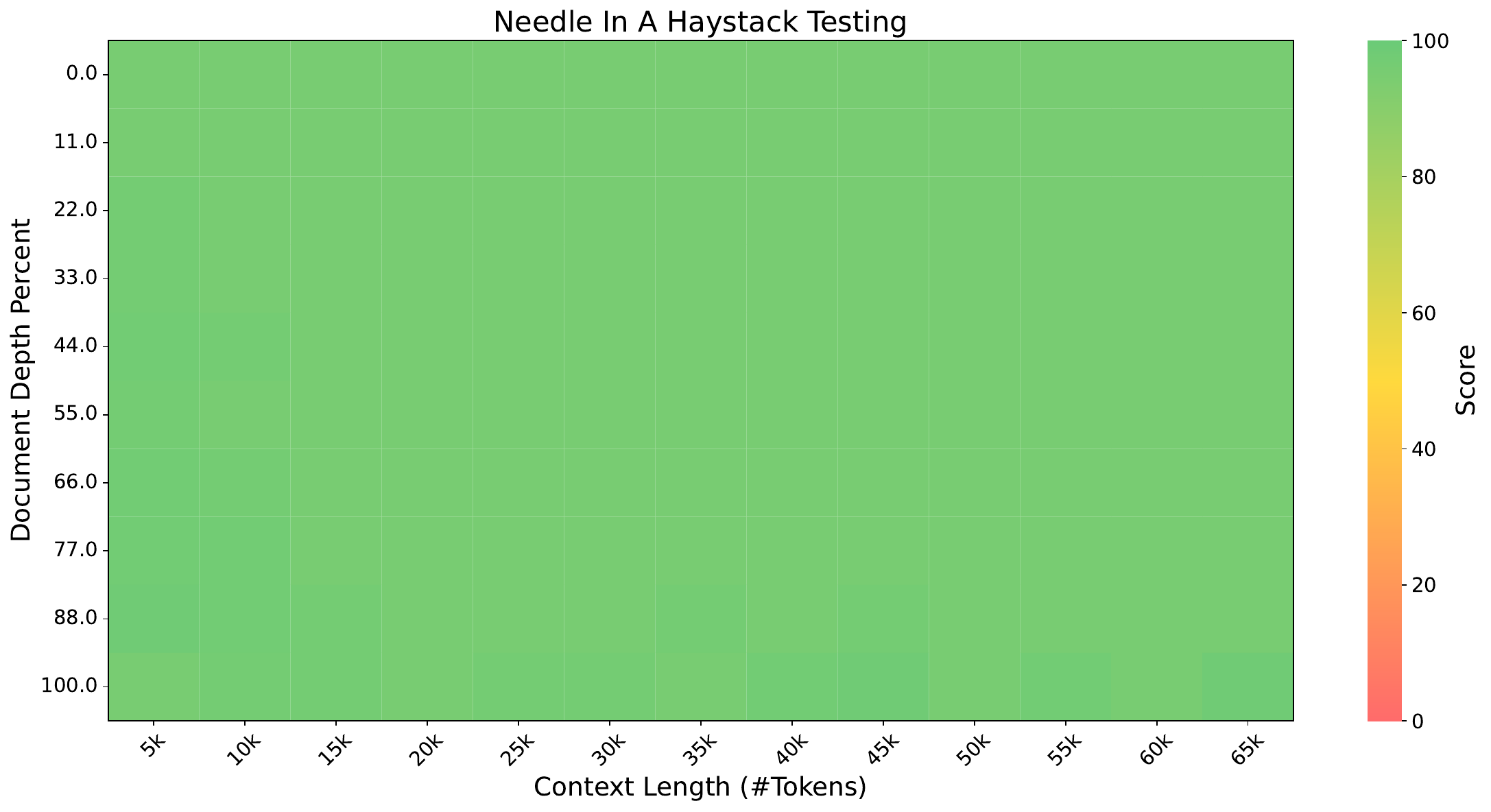"}
	\caption{Needle in A Haystack Testing for Ling-Plus}
	\label{fig:longcontext}
\end{figure}

To address the "lost in the middle" phenomenon \cite{lost}—a common challenge in long-context tasks where key information located in the middle of lengthy documents is often overlooked—we constructed specialized datasets to improve model performance in these scenarios. Specifically, we created single-needle, multi-needle, and multi-hop retrieval datasets by inserting critical information into the middle of documents. These datasets were designed to enhance the model's performance on needle-in-a-haystack tasks and related benchmarks requiring precise identification and retrieval of key information from extended contexts. Figure \ref{fig:longcontext} shows that Ling-Plus achieves nearly perfect performance in "Needle in A Haystack" testing across all context lengths up to 64K.

\subsubsection{Progressive Fine-tuning Strategy} 
To address differences in convergence rates between long-context and short-context tasks, we implemented a two-stage fine-tuning strategy. This approach was designed to ensure robust performance across varying context lengths while preserving the model's foundational capabilities:
\begin{itemize}
    \item \textbf{Short-context adaptation.} In the initial stage, fine-tuning was conducted exclusively on short-context data (contexts with lengths $\leq$4K tokens). This step was aimed at preserving and refining the model's core capabilities, particularly on shorter tasks.
    \item \textbf{Context length extension.} In the second stage, long-context data (contexts ranging from 4K to 16K tokens) was progressively introduced into the fine-tuning process. To balance performance across context lengths, a training sample ratio of 95:5 (short-context to long-context samples) was empirically determined and applied. This ratio was optimized to prevent performance degradation on short-context tasks while enhancing the model's ability to process longer contexts. Additional experiments demonstrated that this progressive fine-tuning approach yields even greater benefits when scaling to longer contexts, such as 16K→64K tokens.
\end{itemize}

\subsubsection{Reinforcement Learning Optimization} 
Consistent with the findings reported in \cite{llama3}, our experiments demonstrated that once the model undergoes successful long-context adaptation during SFT and RL with standard short-context samples effectively improves alignment with human preferences. Notably, this improvement is achieved without degrading the model's performance on long-context tasks. Based on these observations, we adhered to conventional RL training protocols that rely on short-context data.

\subsection{Tool Use}
\label{tool}
In many AI application scenarios, particularly those involving LLM-based agents, the ability to utilize external tools or perform function calls represents a critical capability. To equip the Ling models with this functionality, we train them to interact with the following categories of tools:
\begin{itemize}
    \item \textbf{Public APIs.} Ling models are trained to effectively leverage a wide range of publicly available APIs, such as those provided by RapidAPI \cite{rapidapi2025url}.
    \item \textbf{Application APIs.} Ling models are also trained to utilize application-specific APIs employed by proprietary systems, such as search engine and local service APIs from Alipay agents \cite{alipay2025url}.
    \item \textbf{Synthetic APIs.} Ling models are trained to utilize synthetic APIs generated by our knowledge graph technology.
\end{itemize}

To enhance the tool use ability of our Ling models, we mainly focus on the following two aspects.
\begin{itemize}
    \item \textbf{Synthesis of high-quality tool use data.} (1) \textit{Tool and user instruction collection}:  To enhance the Ling models' ability to interact with diverse tools, we curate a comprehensive dataset comprising open-source APIs from platforms like RapidAPI and GoogleAPI, as well as application-specific APIs for search engines and local services. We employ a knowledge graph technology to design 14 subgraph patterns and their corresponding First-Order Logic (FOL) representations, facilitating the synthesis of APIs and user instructions (queries) and improving the models' generalization in tool use. Furthermore, the dataset is enriched with tool-related user instructions from real-world agent applications and publicly available resources such as ToolBench \cite{qin2023toolllm} and ToolAlpaca \cite{tang2023toolalpaca}, providing a strong foundation for training in tool interaction. (2) \textit{Task planning and system instruction generalization}: Using the knowledge graphs mentioned above, we generate precise tool-calling paths to ensure accuracy and reliability in tool use. To address the variability of tool-based system instructions, we collect established instruction templates, such as LangChain ReACT \cite{langchain2025url}, OpenAI function calling \cite{openai2025functioncalling}, and ModelScope-Agent (Qwen's Agent) \cite{li2023modelscopeagentbuildingcustomizableagent}, and expand them into over 30,000 distinct templates, enabling their application across diverse scenarios and establishing a robust basis for effective task planning and execution.
    \item \textbf{Adaptive tool learning.} To address complex scenarios involving tool use, our Ling models are designed with advanced self-reflection and strategic planning capabilities. The data generation process comprises the following four key components. (1) \textit{Policy agent}: The Ling models serve as policy agents, generating diverse calling responses and leveraging rejected calls to create self-reflective dialogues with the support of reference agents. (2) \textit{Reference agent}: Advanced Ling models or other LLMs are employed to deconstruct user tasks and provide self-reflective feedback when policy models generate error callings. (3) \textit{Quality judgment}: A "model-as-judge" strategy, utilizing advanced Ling models, assigns binary scores to evaluate the success of API calls and overall task completion, ensuring robust and reliable performance.
\end{itemize}
	\section{Results}
\label{section_results}
\subsection{Pre-trained Language Model}

\subsubsection{Evaluation Benchmarks} The Ling base model is pre-trained on multilingual datasets comprising both English and Chinese. Consequently, we evaluate the model's performance on a diverse set of benchmarks that include both Chinese and English. Specifically, the evaluation benchmarks are categorized into the following $4$ types:

\begin{itemize}
    \item \textbf{English.} The English benchmarks contain \emph{multi-subject multiple-choice} task and \emph{language understanding and reading comprehension} task. Multi-subject multiple-choice include MMLU \cite{hendrycks2020measuring}, MMLU-Pro \cite{wang2024mmlu}, MMLU-Redux \cite{gema2024we}. Language understanding and reading comprehension include BBH \cite{suzgun2022challenging}, HellaSwag \cite{zellers2019hellaswag}, PIQA \cite{bisk2020piqa}, ARC challenge \cite{clark2018think}, WinoGrande \cite{sakaguchi2021winogrande}, RACE-Middle and RACE-High \cite{lai2017race}.
    \vspace{4pt}
    \item \textbf{Chinese.} The datasets include C-Eval \cite{huang2023c}, and CMMLU \cite{li2023cmmlu}
    \vspace{4pt}
    \item \textbf{Math.} The datasets include GSM8K \cite{cobbe2021training} and MATH \cite{hendrycks2021measuring}
    \vspace{4pt}
    \item \textbf{Code.} The datasets include HumanEval \cite{chen2021evaluating}, MBPP \cite{austin2021program} and CRUXEval-I and CRUXEval-O \cite{gu2024cruxeval}. 
\end{itemize}

\subsubsection{Benchmarks Optimization}
The evaluation of base LLM models suffer from $2$ critical problems: 
\begin{itemize}
    \item \textbf{Instability in early stage.} Evaluation metrics such as perplexity play an important role in helping us monitor the training of LLM. However, in the early stages of base model pre-training, the model, due to its weak capabilities, shows low differentiation in predicting options in \emph{perplexity-based} evaluations. This leads to fluctuating evaluation results throughout the training process, making it inadequate for monitoring training effectiveness.
    \item \textbf{Lack of instruction-following.} Since the base model lacks instruction-following capabilities, the poor adherence to the answering process or result format negatively affects the evaluation scores, thus failing to accurately reflect the model's true abilities, especially on generation-based evaluation tasks.
\end{itemize}

To tackle the two problems with evaluating base LLM models, we optimize the existing evaluation methods for perplexity-based and generation-based evaluations, to provide results that better match the model's true capability, and increase evaluation stability on LLM.

\paragraph{Optimize Perplexity-Based Evaluation.} To better adapt to the base model's continuation, we change the prediction target from option labels to option content, increasing the differentiation of predictions for each option and improving the trend of capability growth throughout the pre-training process. Details on the optimizations of the above evaluation methods can be found in our corresponding work \cite{luan2025stableconsistentevaluationresults}. 

\begin{figure}[t]
\begin{center}	
\includegraphics[width=0.65\textwidth]{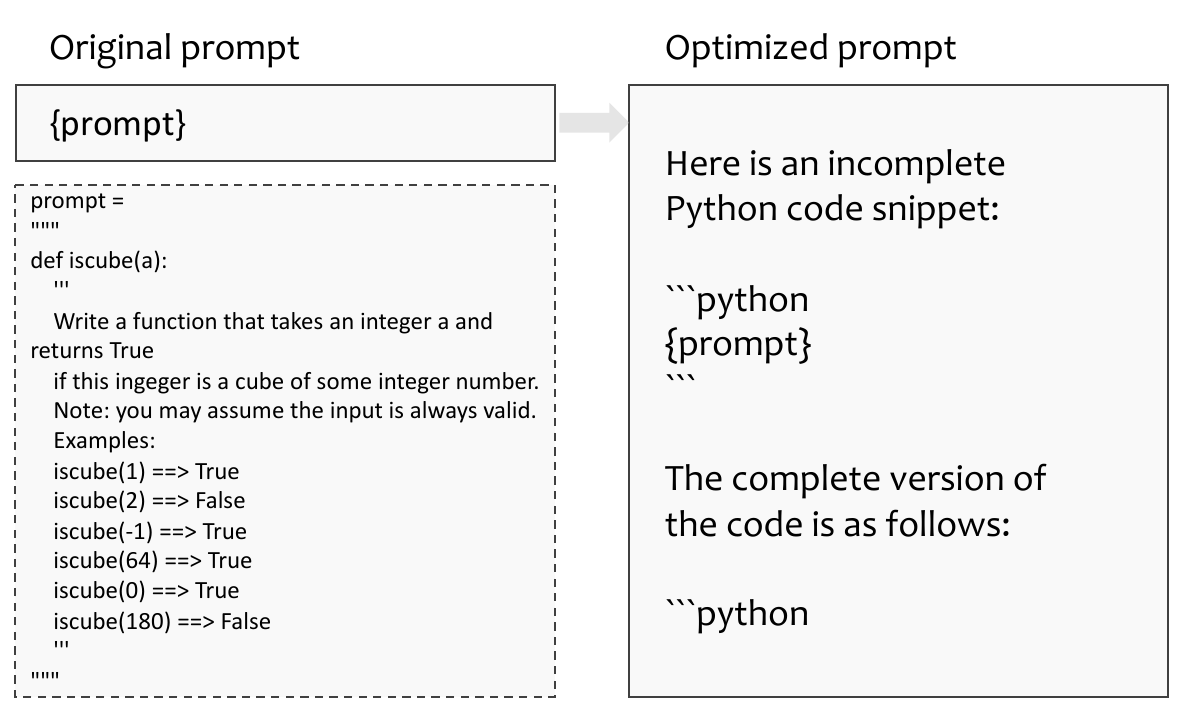}
\caption{An example of optimized prompt for code task.}
\label{fig:code_benchmark_example}
\vspace{4mm}	
\includegraphics[width=1.0\textwidth]{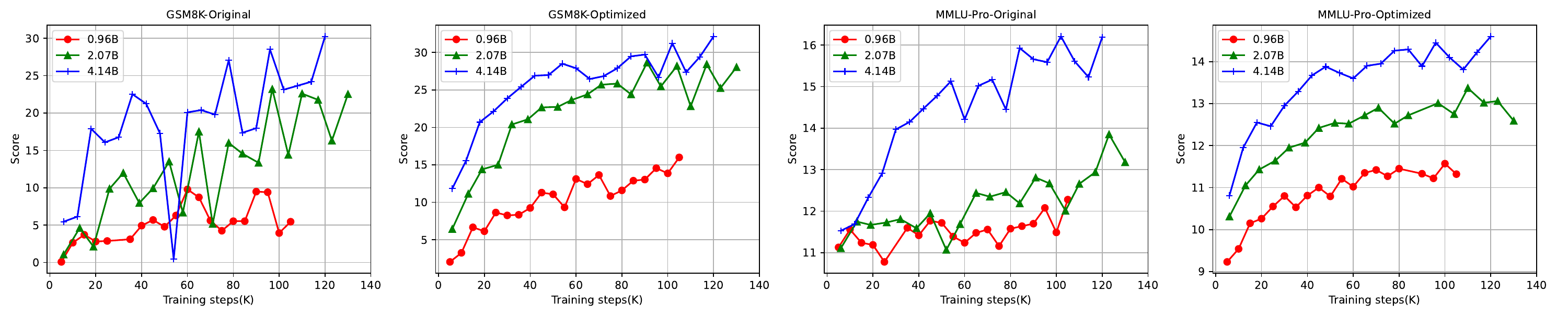}
\caption{Comparisons on improved benchmarks and original benchmarks.}
\label{fig:benchamrk_stability}
\end{center}
\end{figure}

\paragraph{Optimize Generation-Based Evaluation.} To mitigate the impact of the Base model's variable instruction-following capabilities, we optimize the prompt templates to guide its continuation in answering questions. By adding few-shot implicit guidance for reasoning and adherence to format, and configuring stopping criteria to timely conclude reasoning, we improve the effectiveness of question responses, making the evaluation results to better reflect the model's true ability. Taking existing evaluation datasets for \emph{Math} (e.g., GSM8K and MATH) and \emph{Code} (e.g., HumanEval, MBPP, and CRUXE) for example: 

\begin{itemize}
    \item \textbf{Math benchmark.} To better evaluate the capabilities of the base model, we propose several key modifications: providing refined few-shot examples, constructing lightweight prompt templates, and introducing an early stopping mechanism. These improvements can assess the mathematical reasoning capabilities of the base model with more precision.
    \item \textbf{Code benchmark.} We observe that in code tasks, the evaluation of the base model confront two problems: (1) The base model does not understand the actual task requirements. For example, Qwen2-7B-Base do not perform the actual code completion task for 12.19\% of the data in the Humaneval dataset; (2) Since the base model has not aligned with human preferences, it cannot engage in an effective dialogue. Its relatively weak instruction-following capability leads to issues such as truncation and overshooting during the post-processing of code extraction.
\end{itemize}
To address above two issues, we design corresponding solutions: 1) Clearly specify task requirements in the prompt. This enables the base model to clearly understand tasks such as selecting the correct option, calculating the correct answer, or completing the correct code. 2) Provide appropriate prefixes for base model evaluations, this assist the base model in generating the correct continuation and help improve the post-processing extraction of LLM outputs. In Figure \ref{fig:code_benchmark_example} we present an example of our optimized prompt for code task, adding specification and prefixes to original prompt.

\paragraph{Applications in LLM Base Model Training.} To demonstrate the effectiveness of our improved evaluation methods, we compare the changes in evaluation metrics during the early stages of pre-training on several small models under 5B parameters consisting 0.96B, 2.07B and 4.14B models, using our improved benchmarks and the original benchmarks. As shown in Figure \ref{fig:benchamrk_stability}, our improvements in evaluation stability effectively reduce fluctuations in evaluation metrics on the knowledge benchmark MMLU-Pro, and the math benchmark GSM8K, reflecting the stable change in model capabilities as training progresses.

Our optimizations on perplexity-based evaluation and generation-based evaluation, are implemented in the evaluation of both Ling models and other baselines, i.e. DeepSeek, Qwen, LLaMA and Mistral models. 
These optimizations can accurately assess the model's performance in the early stages of training, being used in many application scenarios: providing basis for data ablation experiments, verify the effectiveness of new computing clusters for model training, and facilitate comparisons of training consistency across different computing clusters.

\begin{figure}[t]
\begin{center}	
\includegraphics[width=0.7\textwidth]{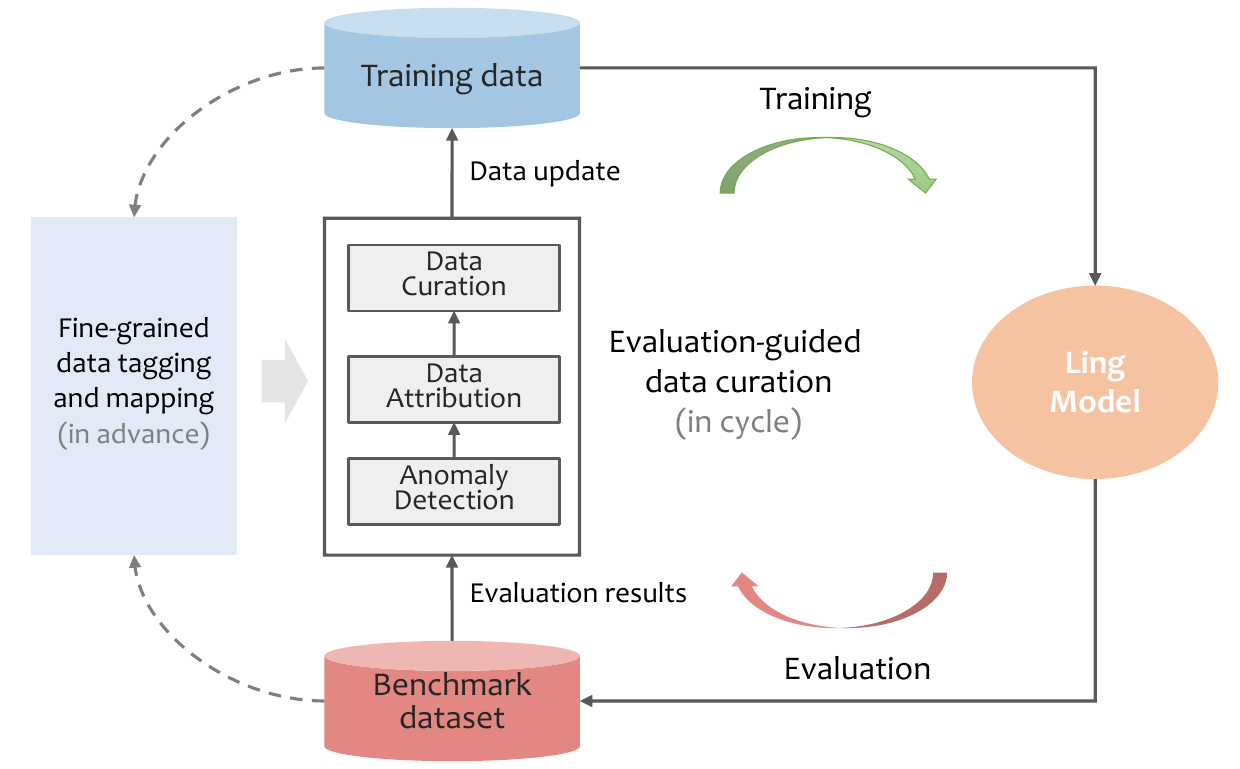}
\caption{Illustration on the process of using evaluations to guide training process.}
\label{fig:linking_evaluation_to_training}
\end{center}
\end{figure}
\paragraph{Linking Evaluations to Training.}
In addition to accurately measuring the model's performance via above optimizations on the evaluation benchmarks, we hope that evaluations of LLM can also help identify issues in the training process, such as problems with the training data. During the evaluation process, we observe that abnormal evaluation results might emerge after the model consumes a certain segment of tokens. This is typically due to problematic data within that segment of training tokens. To identify the reason of such issues and provide real-time feedback for adjustments in the training strategy or training data, we further re-define the ability dimensions corresponding to each evaluation sample within the evaluation benchmark. Simultaneously, we assign the same ability dimensions to the training corpus, enabling the mapping of evaluation results to the training data. This allows us to effectively pinpoint which part of data encounter issues during the training process. We present this whole process in Figure \ref{fig:linking_evaluation_to_training}. 

\FloatBarrier % 插入屏障，强制处理之前的浮动体
\begin{table}[]
    \centering
    \caption{Comparison between Ling-Lite-Base model and other representative models.}
    \vspace{3mm}
\resizebox{0.85\linewidth}{!}{%
\begin{tabular}{clc|c|c|c|c}
\toprule \multicolumn{2}{c}{\multirow{2}{*}{\textbf{Benchmark}\tiny{ (Metric)}}} & \multirow{2}{*}{\textbf{\#shots}} & \textbf{Ling-Lite} & \textbf{Qwen2.5 } & \textbf{LLaMA-3.1 } & \textbf{Mistral-7B}\\
&&& \textbf{-Base} & \textbf{-7B} & \textbf{-8B} & \textbf{-v0.3} \\ \midrule
\multirow{10}{*}{English} & BBH\tiny{ (EM)}  & 3 & 67.38 & 69.07 & 64.02 & 56.12\\
 & MMLU\tiny{ (EM)}  & 5 & 70.88 & 75.50 & 66.61 & 63.45 \\
 & MMLU-Redux\tiny{ (EM)}  & 5 & 65.67 & 70.70 & 60.84 & 58.35\\
 & MMLU-Pro\tiny{ (EM)}  & 5 & 41.47 & 47.60 & 36.72 & 31.47 \\
 & ARC-Challenge\tiny{ (EM)}  & 0 & 87.46 & 91.86 & 81.02 & 72.20 \\
 & WinoGrande\tiny{ (EM)}  & 5 & 74.58 & 75.61 & 77.51 & 77.58\\
 & HellaSwag\tiny{ (EM)}  & 0 & 73.65 & 73.46  & 74.60 & 75.77 \\
 & RACE-Middle\tiny{ (EM)}  & 0 & 89.07 & 91.09 & 90.81 & 71.93 \\
 & RACE-High\tiny{ (EM)}  & 0 & 86.05 & 88.05 & 87.56 & 71.12 \\
 & PIQA\tiny{ (EM)}  & 0 & 78.89 & 79.82 & 80.79 & 81.01 \\
\midrule \multirow{4}{*}{Code} & HumanEval\tiny{ (Pass@1)}  & 0 & 78.66 & 75.00 & 43.97 & 29.88 \\
 & MBPP\tiny{ (Pass@1)}  & 3 & 60.80 & 62.80 & 45.60 & 46.60 \\
 & CRUXEval-I\tiny{ (EM)}  & 1 & 44.38 & 51.38 & 40.88 & 44.00 \\
 & CRUXEval-O\tiny{ (EM)}  & 1 & 44.50 & 48.38 & 36.50 & 34.62 \\
\midrule \multirow{2}{*}{Math} & GSM8K\tiny{ (EM)}  & 4 & 79.68 & 82.71 & 56.56 & 45.94 \\
 & MATH\tiny{ (EM)}  & 4 & 47.48 & 49.42 & 16.94 & 11.26 \\
\midrule \multirow{2}{*}{Chinese} & C-Eval\tiny{ (EM)}  & 5 & 79.33 & 81.14 & 51.50 & 45.86 \\
 & CMMLU\tiny{ (EM)}  & 5 & 80.08 & 81.66 & 52.32 & 44.29 \\
\bottomrule
\end{tabular}
}
    \label{tab:results_ling_lite_base}
\end{table}

\begin{table}[]
\centering
\caption{Comparison between Ling-Plus-Base model and other representative models.}
\vspace{3mm}
\resizebox{0.90\linewidth}{!}{%
    \begin{tabular}{clc|c|c|c|c}
\toprule \multicolumn{2}{c}{\multirow{2}{*}{\textbf{Benchmark}\tiny{ (Metric)}}} & \multirow{2}{*}{\textbf{\#shots}} & \textbf{Ling-Plus} & \textbf{DeepSeek-V2} & \textbf{Qwen2.5} & \textbf{LLaMA-3.1} \\
&&& \textbf{-Base}  & \textbf{-Base} & \textbf{-72B-Base} & \textbf{-70B-Base}\\ \midrule
 \multirow{10}{*}{English} & BBH\tiny{ (EM)}  & 3 & 81.95 & 78.60  & 83.80 & 80.88 \\
 & MMLU\tiny{ (EM)}  & 5 & 81.84 & 79.16  & 86.30 & 79.15 \\
 & MMLU-Redux\tiny{ (EM)}  & 5 & 78.47 & 74.69  & 83.29 & 74.41 \\
 & MMLU-Pro\tiny{ (EM)}  & 5 & 55.18 & 54.17  & 61.40 & 51.42 \\
 & ARC-Challenge\tiny{ (EM)}  & 0 & 92.88 & 90.51  & 96.30 & 91.17 \\
 & WinoGrande\tiny{ (EM)}  & 5 & 77.98 & 84.06  & 81.85 & 84.93 \\
 & HellaSwag\tiny{ (EM)}  & 0 & 77.61 & 80.45  & 80.30 & 79.85 \\
 & RACE-Middle\tiny{ (EM)}  & 0 & 94.15 & 92.41  & 96.20 & 92.48 \\
 & RACE-High\tiny{ (EM)}  & 0 & 92.11 & 90.02  & 93.90 & 88.34 \\
 & PIQA\tiny{ (EM)}  & 0 & 80.09 & 83.35  & 83.80 & 84.06 \\
\midrule \multirow{4}{*}{Code} & HumanEval \tiny{ (Pass@1)}  & 0 & 84.76 & 63.41  & 81.70 & 56.10 \\
 & MBPP\tiny{ (Pass@1)}  & 3 & 71.40 & 66.80  & 76.40 & 66.20 \\
 & CRUXEval-I\tiny{ (EM)}  & 1 & 64.38 & 56.62  & 60.00 & 55.38 \\
 & CRUXEval-O\tiny{ (EM)}  & 1 & 63.75 & 58.75  & 66.12 & 60.62 \\
\midrule \multirow{2}{*}{Math} & GSM8K\tiny{ (EM)}  & 4 & 88.55 & 83.78  & 89.69 & 83.62 \\
 & MATH\tiny{ (EM)}  & 4 & 56.96 & 43.60   & 60.72 & 41.76 \\
\midrule \multirow{2}{*}{Chinese} & C-Eval\tiny{ (EM)}  & 5 & 90.93 & 82.16  & 88.40 & 68.60 \\
 & CMMLU\tiny{ (EM)}  & 5 & 88.56 & 83.01  & 89.50 & 68.84 \\\bottomrule
\end{tabular}
}
    \label{tab:results_ling_plus_base}
\end{table}

\subsubsection{Compared Baselines} 

We release two models of different parameter scales, namely the Ling-Plus model and the Ling-Lite model, and evaluate their performance by comparing them against state-of-the-art open-source models of similar parameter scales, which serve as our baselines. Specifically, the Ling-Plus-Base model is compared with DeepSeek-V2.5, Qwen2.5-72B, and LLaMA-3.1-70B. For Ling-Lite-Base model, we use Qwen2.5-7B, LLaMA-3.1-8B, and Mistral-7B as baselines for its evaluation. The detailed experimental results are listed in Tables \ref{tab:results_ling_lite_base} and \ref{tab:results_ling_plus_base}.

For the evaluation process, we adopt metrics consistent with prior work such as DeepSeek-V2.5 and LLaMA-3.1. Perplexity-based evaluation is employed for datasets including MMLU, MMLU-Redux, MMLU-Pro, HellaSwag, PIQA, WinoGrande, RACE-Middle, RACE-High, ARC-Challenge, C-Eval, and CMMLU. Additionally, generation-based evaluation is used for tasks involving HumanEval, MBPP, CRUXEval, MATH, GSM8K, and BBH.

\subsubsection{Result Analysis}

We compare our Ling pre-trained models with other state-of-the-art open-source base models.
All experiments are conducted using our internal evaluation framework, and we ensure that all models are assessed with same evaluation parameters. In all experiments, we set the temperature of the LLM to 0 and evaluate it in a single run.

\begin{itemize}
    \item \textbf{Ling-Lite.} Comparing our Ling-Lite pre-trained model with other leading 7B+ models. The overall performance of our Ling-Lite-Base model is very close to that of Qwen2.5-7B model, which achieves nearly the best performance across all dimensions we considered. In code and math benchmarks, the Ling-Lite-Base model outperforms Llama3.1-8B and Mistral-7B v0.3. Additionally, in chinese language benchmarks, both the Ling-Lite-Base and Qwen2.5-7B,  which are Chinese open-source models, shows significantly higher scores compared to the other benchmark models.
    \item \textbf{Ling-Plus.} Comparing our Ling-Plus pre-trained model with other leading 70B+models. In the dimensions of code, math, and Chinese language, the overall performance of the Ling-Plus-Base is comparable to that of the Qwen2.5-72B, both models yield similar benchmark scores and higher than those of DeepSeek-V2-Base and Llama3.1-70B-Base. In English language benchmarks, the overall score of the Ling-Plus-Base model is slightly lower than that of Qwen2.5-72B-Base model, but still exceeds the scores of DeepSeek-V2-Base and Llama3.1-70B-Base. It is noteworthy that while Ling-Plus-Base outperforms DeepSeek-V2-Base, it is inferior to its 3.0 version, which currently represents the most advanced open-source model.
\end{itemize}

\subsection{Post-trained Language Model}

\FloatBarrier % 插入屏障，强制处理之前的浮动体
\begin{table}[]
\caption{Comparison between Ling-Lite model and other representative models.}
\centering
\vspace{3mm}
\resizebox{0.85\linewidth}{!}{%
    \centering
\begin{tabular}{cl|c|c|c|c}
\toprule
\multicolumn{2}{c|}{\multirow{2}{*}{\textbf{Benchmark}\tiny{ (Metric)}}} & \textbf{\multirow{2}{*}{Ling-Lite}} & \textbf{Qwen2.5-7B} & \textbf{Llama3.1-8B} & \textbf{Mistral-7B-v0.3}  \\
&&  & \textbf{-Instruct } & \textbf{-Instruct } & \textbf{-Instruct} \\ \midrule
 \multirow{7}{*}{English} & MMLU\tiny{ (EM)} & 71.27 & 74.26 & 68.67 & 61.45 \\
 & MMLU-Redux\tiny{ (EM)} & 70.35 & 75.37 & 67.20 & 35.72 \\
 & MMLU-Pro\tiny{ (EM)} & 49.19 & 55.98 & 47.93 & 18.54 \\
 & IFEval\tiny{ (Prompt Strict)} & 77.99 & 71.16 & 73.01 & 53.45 \\
 & GPQA\tiny{ (Pass@1)}  & 28.66 & 34.47 & 32.80 & 25.63 \\
 & ARC-Challenge\tiny{ (EM)} & 85.08 & 89.15 & 81.69 & 78.98 \\
 & SimpleQA\tiny{ (Correct)} & 4.35 & 5.38 & 15.58 & 4.32 \\
\midrule \multirow{4}{*}{Code} & MultiPL-E\tiny{ (Pass@1)} & 65.78 & 63.11 & 51.66 & 26.27 \\
 & HumanEval\tiny{ (Pass@1)} & 83.54 & 87.20 & 70.73 & 38.41 \\
 & MBPP\tiny{ (Pass@1)} & 64.80 & 61.80 & 59.00 & 40.00 \\
 & LiveCodeBench\tiny{ (Pass@1)} & 15.18 & 16.96 & 11.61 & 8.97\\
\midrule \multirow{3}{*}{Math} & GSM8K\tiny{ (EM)} & 86.88 & 90.60 & 83.02 & 58.61 \\
 & MATH-zero-shot\tiny{ (EM)} & 72.80 & 73.66 & 52.42 & 13.66 \\
 & MATH-few-shot\tiny{ (EM)} & 71.52 & 72.86 & 31.76 & 12.42 \\
 & AIME-2024\tiny{ (Pass@1)} & 6.67 & 16.67 & 0.00 & 0.00 \\
\midrule \multirow{3}{*}{\begin{tabular}{c} Tool \\ Use \end{tabular} } & BFCL-v2\tiny{ (Acc)} & 67.92 & 65.84 & 49.98 & 58.42 \\
 & Nexus\tiny{ (Acc)} & 34.77 & 31.88 & 38.19 & 28.70 \\
 & T-eval\tiny{ (Acc)} & 85.58 & 76.64 & 81.99 & 75.30 \\
\midrule \multirow{3}{*}{Chinese} & C-Eval\tiny{ (EM)} & 73.63 & 78.00 & 53.34 & 43.78 \\
 & CMMLU\tiny{ (EM)} & 72.95 & 78.89 & 53.33 & 42.51 \\
 & C-SimpleQA\tiny{ (Correct)} & 26.07 & 29.63 & 18.94 & 14.10 \\
\midrule \begin{tabular}{c} Open \\ Ended \end{tabular}  & Arena-Hard & 42.09 & 49.20 & 26.94 & 23.47 \\
\bottomrule
\end{tabular}
}
    \label{tab:results_ling_lite}
\end{table}

\begin{table}[t]
\centering
    \caption{Comparison between Ling-Plus and other representative models.}
    \vspace{4mm}
\resizebox{\linewidth}{!}{%
    \centering
\begin{tabular}{cl|c|c|c|c|c|c}
\toprule \multicolumn{2}{c|}{\multirow{2}{*}{\textbf{Benchmark}\tiny{ (Metric)}}} & \textbf{Ling-Plus} & \textbf{Ling-Plus} & \textbf{DeepSeek-V2.5} & \textbf{Qwen2.5-72B} & \textbf{Llama3.1-70B} & \multirow{2}{*}{\textbf{GPT4o-0806}} \\
 & & \textbf{\tiny{(Device-A accelerator)}} & \textbf{\tiny{(Device-D accelerator)}} & \textbf{-1210-Chat} & \textbf{-Instruct} & \textbf{-Instruct} & \\
\midrule \multirow{7}{*}{English} & MMLU\tiny{ (EM)} & 82.33 & 82.52 & 80.74  & 84.30 & 81.68 & 86.46\\
 & MMLU-Redux\tiny{ (EM)} &  83.90 & 83.95 & 81.25  & 85.56 & 80.48 & 88.00\\
 & MMLU-Pro\tiny{ (EM)} &  67.57 & 67.92 & 64.47  & 70.77 & 66.94 & 74.83\\
 & IFEval\tiny{ (Prompt Strict)} & 83.73 & 85.65 & 79.67  & 82.44 & 82.44 & 86.17 \\
 & GPQA\tiny{ (Pass@1)} &  43.81 & 42.55 & 41.67  & 47.98 & 42.42 & 52.53 \\
 & ARC-Challenge\tiny{ (EM)} & 93.90 & 94.24 & 92.88  & 95.25 & 93.22 & 95.25 \\
 & SimpleQA\tiny{ (Correct)} & 11.86 & 11.93 & 10.91  & 12.31 & 10.10 & 40.07 \\
\midrule \multirow{4}{*}{Code} & MultiPL-E\tiny{ (Pass@1)} & 69.79 & 69.39 & 71.04  & 69.08 & 61.32 & 69.97 \\
 & HumanEval\tiny{ (Pass@1)} & 90.24 & 89.02 & 88.41  & 88.41 & 79.88 & 91.46 \\
 & MBPP\tiny{ (Pass@1)} &  76.60 & 76.60 & 78.80 & 78.40 & 72.80 & 80.20 \\
 & LiveCodeBench\tiny{ (Pass@1)} &  26.79 & 25.89 & 31.25  & 26.79 & 12.50 & 34.20 \\
\midrule \multirow{4}{*}{Math} & GSM8K\tiny{ (EM)} & 94.47 & 94.16 & 90.67  & 93.40 & 92.12 & 96.21 \\
 & MATH-zero-shot\tiny{ (EM)} &  78.82 & 78.76 & 76.94  & 81.14 & 57.86 & 77.94 \\
 & MATH-few-shot\tiny{ (EM)} &  78.63 & 78.57 & 74.39  & 80.46 & 52.46 & 75.34 \\
 & AIME-2024\tiny{ (Pass@1)} & 33.33  & 26.67 & 23.33  & 20.00 & 23.33 & 20.00 \\
\midrule \multirow{3}{*}{\begin{tabular}{c} Tool \\ Use \end{tabular} } & BFCL-v2\tiny{ (Acc)} & 74.90 & 75.65 & 58.24  & 73.39 & 60.51 & 62.19 \\
 & Nexus\tiny{ (Acc)} & 50.10 & 50.09 & 45.75  & 51.99 & 52.07 & 51.55 \\
 & T-eval\tiny{ (Acc)} & 89.25 & 89.14 & 75.37  & 87.62 & 86.29 & 88.44 \\
\midrule \multirow{3}{*}{Chinese} & C-Eval\tiny{ (EM)} & 86.87 & 86.55 & 82.25  & 88.02 & 68.25 & 77.29 \\
 & CMMLU\tiny{ (EM)} & 86.59 & 86.49 & 81.19  & 87.44 & 70.92 & 80.04 \\
 & C-SimpleQA\tiny{ (Correct)} & 51.77 & 52.13 & 57.40  & 50.93 & 40.69 & 61.43 \\
\midrule Open & \multirow{2}{*}{Arena-Hard} & \multirow{2}{*}{74.25} & \multirow{2}{*}{74.56} & \multirow{2}{*}{77.92} & \multirow{2}{*}{78.98} & \multirow{2}{*}{58.46} & \multirow{2}{*}{80.40} \\
 Ended &  &  &    &   &   &   &     \\
\bottomrule
\end{tabular}
}
    \label{tab:results_ling_plus}
\end{table}

\subsubsection{Evaluation Benchmarks} In addition to the benchmarks used for evaluating the base model, we introduce additional benchmarks to assess the capabilities of the instructed model in English and Chinese on \emph{language understanding and reading comprehension} task, \emph{Code} and \emph{Math} task. Specifically, for English language datasets, we incorporate IFEval\cite{zhou2023instruction}, GPQA-Diamond \cite{rein2024gpqa}, and SimpleQA \cite{simpleqa2024openai}; for Chinese language datasets, we add C-SimpleQA \cite{he2024chinese}; for \emph{Code} task, we use MultiPL-E \cite{cassano2022multipl} \footnote{Excluding C\#, due to the coding environment.} and LiveCodeBench \cite{jain2024livecodebench};
for \emph{Math} tasks, we add AIME \cite{aime}.

Additionally, to further explore the capability of the model serving as agents, and simulate the real-world applications, we supplement $2$ categories of benchmarks focusing on \emph{tool use} task and \emph{open-ended generation} task, to further evaluate the chat model's ability. The \emph{tool use} benchmarks include BFCL \cite{berkeley_function_calling_leaderboard}, Nexus \cite{srinivasan2023nexusraven} and T-eval \cite{chen2023t}, and the \emph{open-ended generation} use Arena-Hard \cite{li2024crowdsourced}. 

\subsubsection{Baseline Comparison}
Similar to the baselines used for evaluating the base model, for chat models with different parameter scales, we adopt instructed models of corresponding scales as baselines. We compare the Ling-Lite model to Qwen2.5-7B-Instruct, Llama3.1-8B-Instruct and Mistral-7B-v0.3-Instruct in Table \ref{tab:results_ling_lite}. We compare our Ling-Plus model to DeepSeek-V2.5-Chat, Qwen2.5-72B-Instruct and Llama3.1-70B-Instruct in Table \ref{tab:results_ling_plus}. 

\subsubsection{Result Analysis} 
Comparing the Ling-Plus model and the Ling-Lite model to the baselines,  considering $5$ tasks including \emph{Language understanding and reading comprehension} (both English and Chinese), \emph{Code}, \emph{Math}, \emph{Tool use} and \emph{Open-ended Generation}, we have the following findings: 
\begin{itemize}

\item \textbf{English \& Chinese language understanding.} MMLU is a widely used LLM benchmark across knowledge domains and tasks. The Ling-Lite demonstrates performance comparable to Qwen2.5-7B-Instruct, while outperforming Llama3.1-8B-Instruct and Mistral-7B-v0.3-Instruct. Ling-Plus achieve performance comparable to DeepSeek-V2.5-Chat and Qwen2.5-72B-Instruct. On GPQA dataset, Ling-Plus is comparable to DeepSeek-V2.5 and Ling-Lite is comparable to Mistral-7B-v0.3. On the instruction-following benchmark IFEval, Ling-Lite achieves the best performance compared to other small-size baselines, and Ling-Plus is also comparable to other large-size baseline models. ARC-challange is a more difficult subset of ARC. Both our Lite and Plus models maintain performance comparable to other baselines. On the factual knowledge benchmark SimpleQA, all models exhibit relatively poor performance, our Ling-Plus has a similar performance compared to DeepSeek-V2.5.

On Chinese benchmarks, as Qwen, Deepseek and our Ling model are trained with more Chinese language data, they demonstrate significantly superior performance compared to Llama and Mistral. Both our Lite and Plus performs slightly better than Deepseek, and is comparable to Qwen on CEval and CMMLU, while Deepseek performs better on C-SimpleQA.

\item \textbf{Math \& code.} On math and code benchmarks, Ling-Lite demonstrates performance comparable to Qwen2.5-7B, while both Qwen and Ling-Lite outperforms Llama3.1-8B and Mistral-7B-v0.3. Ling-Plus model exhibits performance better than DeepSeek-V2.5, closely approximating Qwen2.5-72B. 

\item \textbf{Tool use}. Tool use is an important and challenging task for LLMs. The tool use capability enables LLMs to work as agents, control robotic system and integrate with many software tools. Compared with other baseline models, in most cases, both our Ling-Plus and Ling-Lite achieve the best performance on tool use benchmarks, especially on BFCL-v2 and T-eval. On the Nexus dataset, our model achieve comparable performance to other baselines, with 4 points lower than Llama3.1-8B. As an open-source model, we hope our Ling models, including Ling-Plus and Ling-Light, can provide some insights for the community, facilitating the deployment of LLMs as agents capable of handling more complex tasks. 

Furthermore, we observe that using few-shot prompts, can have negative impacts on the model's performance. On MATH benchmark, the zero-shot version demonstrate significantly superior performance compared to few-shot version. This suggests that caution should be exercised when employing the few-shot setting with instruct models, on Math tasks.

\item \textbf{Open-ended generation.} On open-ended benchmark Arena-Hard, which consists of difficult code and mathematical problems, our Ling-Lite model outperforms Llama3.1-8B and Mistral-7B-v0.3, and our Ling-Plus model demonstrates comparable performance to DeepSeek-V2.5.

\item \textbf{Consistency on different AI accelerator.}
Last but not least, we compare the performance of Ling-Plus model using different AI accelerators, i.e., Device-A AI accelerator and Device-D AI accelerator, on various benchmarks. As shown in Table \ref{tab:results_ling_plus}, the Ling-Plus model achieve almost identical results on each benchmark regardless of which AI accelerator is used.

\end{itemize}

\subsection{Safety}

\begin{table}[]
\caption{Safety performance comparison between Ling-Lite model and other baseline models.}
\centering
\vspace{3mm}
\resizebox{0.8\linewidth}{!}{%
    \centering
\begin{tabular}{cl|c|c|c|c}
\toprule \multicolumn{2}{c|}{\multirow{2}{*}{\textbf{Benchmark}}} & \multirow{2}{*}{\textbf{Ling-Plus}} & \textbf{DeepSeek-V2.5}    & \textbf{Qwen2.5-72B}  & \textbf{Llama3.1-70B}  \\
& & & \textbf{-1210-Chat}  & \textbf{-Instruct} & \textbf{-Instruct} \\
\midrule \multirow{2}{*}{safety} & Arena Safety & 89.50 & 75.50  & 92.50 & 80.50 \\
 & Cvalues & 96.09 & 96.26  & 96.26 & 93.52 \\
\midrule \multirow{2}{*}{false refusal} & Xstest & 98.40 & 97.20  & 98.80 & 100.00 \\
 & Orbench-Hard-1k & 90.24 & 91.96  & 77.15 & 60.11 \\
\midrule \multicolumn{2}{c|}{average score} & 93.56 & 90.23  & 91.18 & 83.53 \\
\bottomrule
\end{tabular}
}
    \label{tab:results_ling_safety}
\end{table}

\subsubsection{Evaluation Benchamarks}

In addition to evaluating Ling models ability on various evaluation benchmarks, we also assess the models' safety performance. Two evaluation datasets are constructed from the open-sourced data: 1) Arena Safety is constructed by randomly sampling 803 questions from a subset of lmsys-chat-1m \cite{zheng2023lmsys} which is identified as risk by OpenAI moderation API \cite{markov2023holistic}, and the responses are evaluated by Llama-Guard3 \cite{llamaguard3url}; 2) Cvalues \cite{xu2023cvalues} uses 1711 multiple-choice questions to assess responsibility in a chinese context.

Moreover, previous work \cite{rottger2023xstest} found that improving LLM's harmlessness can lead to a decrease in helpfulness. To balance this trade-off, we additionally introduce two over-refusal evaluation benchmarks: (1) \textit{Xstest} \cite{rottger2023xstest} contains 250 non-risky but easily erroneously refused questions; (2) \textit{Orbench-Hard-1k} \cite{cui2024or} includes 1000 more challenging questions for a large-scale over-refusal test. Both Xstest and Orbench-Hard-1k use GPT-4o to judge if the LLM refuses to answer.

In Table \ref{tab:results_ling_safety}, we present the safety and false refusal results by comparing the Ling models to baseline models. The safety metric reflects the proportion of safe responses, and the false rejection metric reflects the proportion of non-refusal responses. Both metrics are the higher the better.

\subsubsection{Results analysis}

As in Table \ref{tab:results_ling_safety}, both Ling-Plus and Qwen2.5-72B-Instruct stand out in terms of safety, and Ling-Plus performs better considering false refusal. The DeepSeek series models exhibit the least false refusal phenomenon, but they show lower safety on risk questions within the lmsys-chat-1m. Ling-Plus demonstrates a better overall trade-off between safety and refusal, achieving the best results in terms of the average of these metrics.
	\section{Bitter Lessons}\label{bitter_lessons}
Training LLM is a challenging and resource-intensive process, often accompanied by various technical difficulties. Errors and exceptions are common, with some being relatively straightforward to resolve while others require significant time and effort. To aid researchers and practitioners in this field, we have compiled a summary of frequent issues encountered during training, along with strategies to address them.

\subsection{Training Stability}
Training stability encompasses challenges such as loss spikes, loss divergence, and expert load imbalance, particularly in MoE models. These issues can hinder performance or even lead to training failure. Based on empirical observations:
\begin{itemize}
    \item \textbf{Loss spikes.} Loss spikes are abrupt increases in training loss and are often caused by specific data and optimizer state combinations. Narrow, sharp spikes tend to have a minimal impact on performance, whereas wide, prolonged spikes can adversely affect both stability and model performance. During MoE training, such spikes may also result from hardware issues, such as malfunctioning accelerators or under-performing compute nodes. To mitigate these effects, we implemented a series of measures, including retry and skip mechanisms. When wide spikes occur for the first time, the affected update is skipped, the data is saved, and the training step is retried, as is described in Section\ref{sec:loss_spikes}. If spikes persist upon retrying, we automatically reduce the learning rate during the affected step. This strategy has proven relatively effective in minimizing the impact of loss spikes compared to leaving them unaddressed.
    \item \textbf{Loss divergence.} Loss divergence, which halts training progress, is often caused by numerical instabilities in softmax layers. To counter this, our training architecture incorporates two mitigation techniques: (1) the use of HeadNorm to stabilize the softmax layer in the language modeling head and (2) the application of zloss to the router softmax layer for expert routing. These methods are inspired by the work of Zoph et al. on designing stable MoE architectures \cite{zoph2022stmoedesigningstabletransferable}.
    \item \textbf{Expert load imbalance.} Maintaining balanced expert utilization is essential for the effectiveness of MoE models. Wide loss spikes can significantly disrupt expert load balance by causing abrupt gradient surges, which destabilize the routing equilibrium. Once experts become imbalanced, the issue tends to escalate, leading to widespread instability across the model. By integrating our spike mitigation techniques with balance loss and the aforementioned router zloss, we successfully achieved stable training for an MoE model containing hundreds of billions of parameters. This approach resulted in a stable loss trajectory, with no observed instances of loss divergence, wide loss spikes, or disruptions in expert routing balance.
\end{itemize}

\subsection{Cross-Platform Alignment}
The migration of LLMs training across different platforms presents a multifaceted challenge, primarily due to discrepancies in the implementation of fundamental operations and framework-level distinctions. These variations can lead to divergent training outcomes, underscoring the necessity for rigorous alignment strategies. To facilitate the migration of Ling—a large-scale LLM—to multiple platforms, we conducted extensive preparatory experiments aimed at ensuring the consistency of basic operations and communication algorithms across platforms, while accounting for minor precision errors inherent to numerical computations. Only after successful validation of these foundational components did we proceed to large-scale LLM training.

However, validating basic operations alone proved insufficient for achieving seamless cross-platform migration. During subsequent training phases, significant disparities in loss convergence were observed between platforms post-migration. To address this issue, we extended our alignment efforts beyond basic operations to encompass the frameworks themselves. This process required the elimination of all potential sources of divergence; otherwise, pinpointing the root cause of errors would have been infeasible. Consequently, we achieved full alignment of fundamental operations, including matrix multiplication (\texttt{matmul}) and linear transformations, across both platforms.
\begin{figure}[t]
    \begin{center}	
    \includegraphics[width=0.9\textwidth]{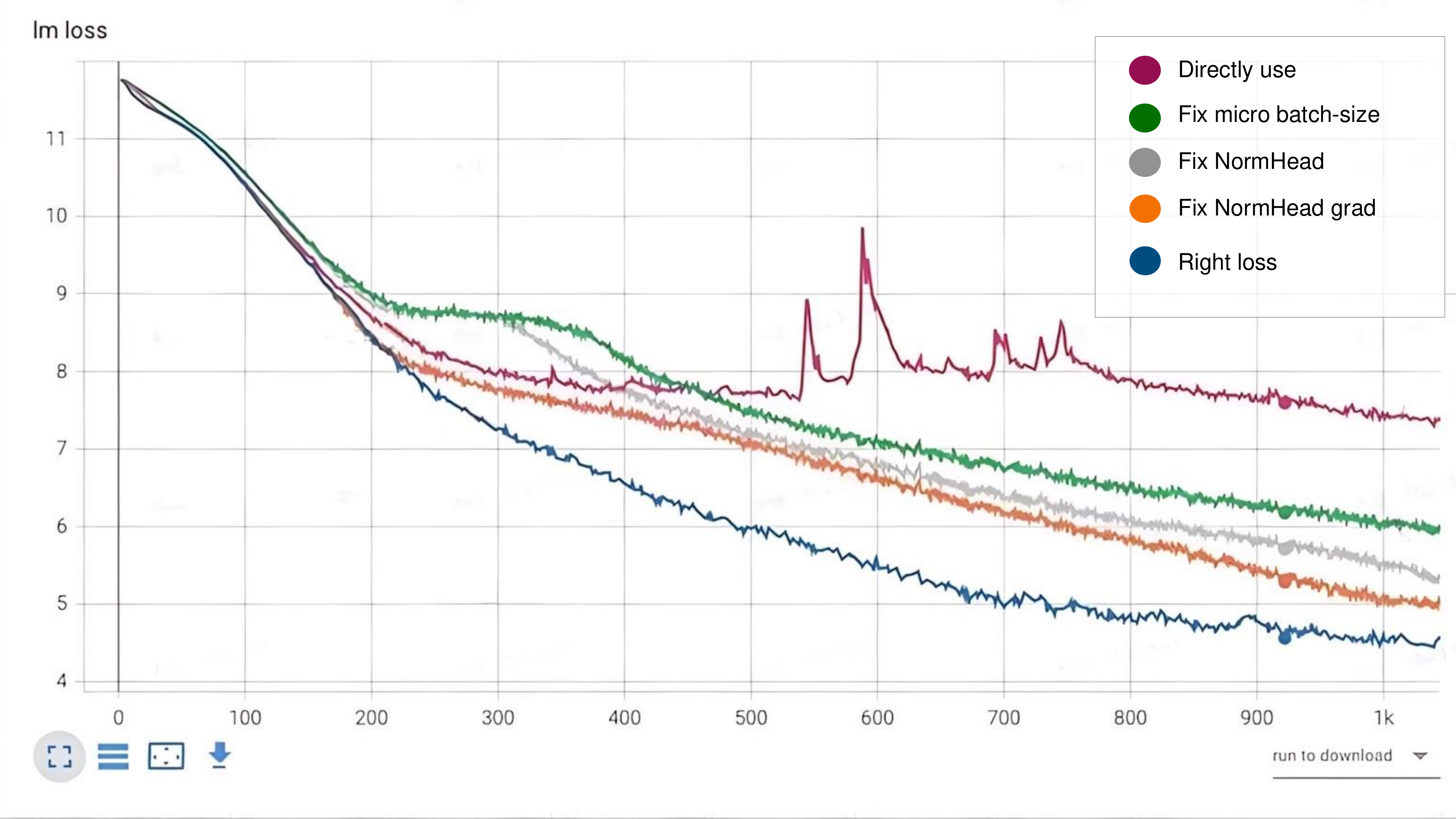}
    \caption{When switching between different hardware platforms, even after verifying the consistency among various operators, it is still necessary to examine the detailed operations and communication behaviors within the frameworks to ensure that the final results meet the expected outcomes. This is our record of fixing the loss curve in the Megatron vendor version on Device A. }
    \label{fig:linglite-diagnosis}
    \end{center}
    \end{figure}
At the framework level, discrepancies in the implementation of modules—such as Attention mechanisms, Multi-Layer Perceptrons (MLPs), and Router components—were addressed to avoid precision errors stemming from floating-point arithmetic. This effort resulted in complete alignment of forward-pass computations across platforms. In this process, we resolved issues arising from variations in tensor parallelism (TP) and auxiliary loss calculations and corrected errors in certain communication operations. During backward-pass computations, leveraging the insights gained from aligning the forward pass allowed us to efficiently identify and rectify errors in gradient propagation, particularly in the router components. While such issues may appear negligible in isolation or during unit testing, their cumulative effect over the course of training can significantly impact convergence outcomes for LLMs. Even minor discrepancies, when compounded over many iterations, can lead to substantial deviations in final loss convergence.

Thus, achieving full alignment of both forward and backward computational passes is imperative for training on new platforms or frameworks. This process not only ensures training correctness and stability but also enhances the understanding of platform-specific characteristics. Furthermore, it facilitates the development of new features and optimization strategies, contributing to future advancements in LLM performance and scalability. Our repair process can be referenced in Figure~\ref{fig:linglite-diagnosis}.

	\section{Conclusion}
This report has addressed the challenges associated with training large-scale MoE models, including cost inefficiency and resource limitations, by proposing innovative strategies to improve efficiency in resource-constrained environments. Specifically, we introduced two open-source MoE models, Ling-Lite and Ling-Plus, which are designed to reduce training costs through advancements in architectural design, frameworks, and storage optimization. Our experimental findings have demonstrated that a 300B MoE LLM can be effectively trained on lower-performance devices while achieving comparable performance to similar scale of dense and MoE models, such as Qwen2.5-72B-Instruct and DeepSeek-V2.5-1210-Chat. Also, compared with the high-performance devices, utilizing a lower-specification hardware system during the pre-training phase has demonstrated significant cost savings, reducing computing cost by approximately 20\%. In this report, we also presented our comprehensive optimization solutions for model training across diverse computational resources. These include improvements to model architecture and training strategies, enhancements to training anomaly handling mechanisms, optimization of model evaluation processes, and advancements in the ability of tool use. To continue the development of the Ling series of LLMs, we plan to release our coder model in the near future.
	\section{Authors}
Binwei Zeng, Chao Huang, Chao Zhang, Changxin Tian, Cong Chen, Dingnan Jin, Feng Yu, Feng Zhu, Feng Yuan, Fakang Wang, Gangshan Wang, Guangyao Zhai, Haitao Zhang, Huizhong Li, Jun Zhou, Jia Liu, Junpeng Fang, Junjie Ou, Jun Hu, Ji Luo, Ji Zhang, Jian Liu, Jian Sha, Jianxue Qian, Jiewei Wu, Junping Zhao, Jianguo Li, Jubao Feng, Jingchao Di, Junming Xu, Jinghua Yao, Kuan Xu, Kewei Du, Longfei Li, Lei Liang, Lu Yu, Li Tang, Lin Ju, Peng Xu, Qing Cui, Song Liu, Shicheng Li, Shun Song, Song Yan, Tengwei Cai, Tianyi Chen, Ting Guo, Ting Huang, Tao Feng, Tao Wu, Wei Wu, Xiaolu Zhang, Xueming Yang, Xin Zhao, Xiaobo Hu, Xin Lin, Yao Zhao, Yilong Wang, Yongzhen Guo, Yuanyuan Wang, Yue Yang, Yang Cao, Yuhao Fu, Yi Xiong, Yanzhe Li, Zhe Li, Zhiqiang Zhang, Ziqi Liu, Zhaoxin Huan, Zujie Wen, Zhenhang Sun, Zhuoxuan Du, and Zhengyu He.
	\bibliographystyle{plainnat}
	\bibliography{btr}
\end{document}